\documentclass[fleqn,10pt]{wlscirep}
\usepackage[utf8]{inputenc}
\usepackage[T1]{fontenc}
\usepackage{amsmath,amssymb,amsfonts}
\usepackage{multirow}
\usepackage{subcaption}
\usepackage{algorithm}
\usepackage{algorithmic}

\title{BioVessel-Net and RetinaMix: Unsupervised Retinal Vessel Segmentation from OCTA Images}

\author[1,2,3,+]{Cheng Huang}
\author[2,4,+]{Weizheng Xie}
\author[1]{Fan Gao}
\author[1]{Yutong Liu}
\author[1,5]{Ruoling Wu}
\author[2,6]{Zeyu Han}
\author[6]{Jingxi Qiu}
\author[1,$\dag$]{Xiangxiang Wang}
\author[1,7,$\dag$]{Zhenglin Yang}
\author[1,8]{Hao Wang}
\author[1,$\dag$]{Yongbin Yu}

\affil[1]{University of Electronic Science and Technology of China, Chengdu, 610056, China}
\affil[2]{Southern Methodist University, Dallas, 75205, USA}
\affil[3]{University of Texas Southwestern Medical Center, Dallas, 75390, USA}
\affil[4]{Karat by Lollicup, Rockwall, 75032, USA}
\affil[5]{University Medical Center Utrecht, Utrecht, 3584 CX, The Netherlands}
\affil[6]{Georgetown University, Washington, 20057, USA}
\affil[7]{Sichuan Provincial People's Hospital, Chengdu, 610072, China}
\affil[8]{University of Connecticut, Storrs, 06269, USA}

\affil[$\dag$]{Corresponding Author: ybyu@uestc.edu.cn}
\affil[+]{these authors contributed equally to this work}

\begin{abstract}
Structural changes in retinal blood vessels are critical biomarkers for the onset and progression of glaucoma and other ocular diseases. However, current vessel segmentation approaches largely rely on supervised learning and extensive manual annotations, which are costly, error-prone, and difficult to obtain in optical coherence tomography angiography. Here we present BioVessel-Net, an unsupervised generative framework that integrates vessel biostatistics with adversarial refinement and a radius-guided segmentation strategy. Unlike pixel-based methods, BioVessel-Net directly models vascular structures with biostatistical coherence, achieving accurate and explainable vessel extraction without labeled data or high-performance computing. To support training and evaluation, we introduce RetinaMix, a new benchmark dataset of 2D and 3D OCTA images with high-resolution vessel details from diverse populations. Experimental results demonstrate that BioVessel-Net achieves near-perfect segmentation accuracy across RetinaMix and existing datasets, substantially outperforming state-of-the-art supervised and semi-supervised methods. Together, BioVessel-Net and RetinaMix provide a label-free, computationally efficient, and clinically interpretable solution for retinal vessel analysis, with broad potential for glaucoma monitoring, blood flow modeling, and progression prediction. Code and dataset are available: \color{blue}{\url{https://github.com/VikiXie/SatMar8}}.
\end{abstract}

\begin{document}

%\flushbottom
\maketitle

%\thispagestyle{empty}

%\noindent\textbf{Keywords} Abbreviations should be introduced at the first mention in the main text – no abbreviations lists. Suggested structure of main text (not enforced) is provided below.

\section*{Introduction}

Glaucoma is a leading cause of irreversible blindness, often progressing silently until substantial vision loss has occurred\cite{fairdomain,fairseg}. Recent evidence suggests that choroidal microvasculature dropout may serve as a potential biomarker for disease progression\cite{glaboost-2}. Optical Coherence Tomography Angiography (OCTA) enables non-invasive, depth-resolved visualization of these vascular alterations, yet reliable quantification of choroidal vessel density remains a significant challenge\cite{octa500,xgan-1}.

Current imaging analytics for glaucoma primarily rely on computer vision techniques\cite{lag-1,lag-2,LbNets,cnn-1,convit,v11,CauSSL,gan,cyclegan}. In particular, supervised learning segmentation models (SLSMs)\cite{unet,unetpp,attunet,maskrcnn,MedSAM,Swin-UNet,TransUNet,skin-4,skin-5} have become widely used. However, SLSMs are often ill-suited for choroidal vessel analysis in OCTA due to their strong dependence on large annotated datasets\cite{fairclip,fairvision}, which are extremely difficult to obtain. The complex morphology of OCTA images, including irregular vessel patterns, dense and intersecting capillary networks, and the presence of large vessels that must be excluded, further complicates manual labeling and automated segmentation\cite{octa500}.

Emerging studies have shown that retinal blood vessels follow biostatistical laws\cite{refuge,age}, where vessel radius variations along branching structures suggest that segmentation can bypass purely pixel-based mapping by leveraging radius as a discriminative feature\cite{bg-4,bg-5,bg-6,glalstm-3}. Motivated by this observation, we propose BioVessel-Net, an unsupervised framework that integrates generative adversarial networks with retinal vessel biostatistics to achieve precise main-vessel segmentation from OCTA images without requiring labeled data. To further address data scarcity, we introduce RetinaMix, a benchmark dataset of high-resolution 2D and 3D OCTA images with clear capillary distribution, supporting vessel segmentation, 3D reconstruction, and blood flow analysis.

This paper substantially extends our preliminary work presented at ICONIP 2025 (oral presentation; acceptance rate 6\%), where we first introduced the X-GAN framework for unsupervised main-vessel segmentation\cite{xgan-1}. In this journal version, we present BioVessel-Net, which incorporates vessel biostatistical priors to enhance clinical interpretability. We further contribute RetinaMix, a curated dataset with demographic diversity, and conduct comprehensive evaluations including cross-dataset validation, ablation and sensitivity analyses, and efficiency benchmarking. These extensions transform the prototype into a clinically generalizable system with strong reproducibility, aligning with the translational scope of \textit{Scientific Reports}.

This paper makes the following major contributions:

\begin{itemize}
    \item We propose the BioVessel-Net, a biostatistics-guided unsupervised framework for retinal vessel analysis, which integrates vascular radius priors into a GAN-enhanced DFS segmentation pipeline. This design achieves near-perfect accuracy while maintaining strong biological interpretability.
    
    \item We construct and release the RetinaMix, a new high-resolution 2D/3D OCTA dataset comprising 550 images from 250 subjects, with balanced demographic representation across gender and race. This dataset provides an open benchmark for vessel structure analysis and glaucoma screening.
    
    \item We perform comprehensive evaluations across multiple datasets (RetinaMix, OCTA-500\cite{octa500}, ROSE\cite{ROSE}), including cross-dataset generalization, hyperparameter sensitivity, ablation studies, and efficiency analysis against SOTA supervised models. Results demonstrate both superior accuracy and computational efficiency.
    
    \item We demonstrate that the proposed threshold choice ($R_{min}\approx0.2$) is consistent with independent ophthalmological findings on capillary coverage ($\approx$18-26\%), thereby validating the biological plausibility of our approach and reinforcing its potential for clinical translation.
\end{itemize}

\section*{Related Work}

\subsection*{Public Resources for Glaucoma Research}

Glaucoma-related research has benefited from a wide range of publicly available fundus photograph datasets. Among the most frequently used are Drishti-GS\cite{drishti} with 101 color fundus images that include expert annotations of the optic disc and cup, ORIGA\cite{origa} with 650 images that provide optic nerve head labels and cup-to-disc ratio measurements, and RIM-ONE\cite{rimone} with 169 images collected from normal, glaucoma, and ocular hypertension cases. Population-based resources such as the Singapore Chinese Eye Study\cite{sces}, which involves more than 3,300 subjects with comprehensive clinical annotations, offer additional large-scale epidemiological support. More recent contributions include MURED\cite{mured} with 2208 images and ROSE\cite{ROSE} with 229 images, both of which provide curated fundus photographs with detailed labels for glaucoma assessment. Classic benchmarks for vessel segmentation such as DRIVE\cite{drive} with 40 images and STARE\cite{stare} with 20 images are often combined with glaucoma datasets to evaluate vascular biomarkers that are relevant to disease progression.

In addition to fundus photographs, OCTA-based datasets have become important for capturing microvascular changes associated with glaucoma. Notable examples include OCTA-500\cite{octa500}, which contains imaging from 500 subjects, and the large-scale Harvard datasets, namely Harvard-GDP\cite{gdp} with 21,059 samples and Harvard-GF\cite{fairness} with 3300 samples, which provide optical coherence tomography angiography linked to glaucoma phenotypes. Beyond single-modality imaging, multimodal and fairness-oriented datasets have also been introduced. FairSeg\cite{fairseg} and FairDomain\cite{fairdomain} each contain 10,000 images or subjects, while FairCLIP\cite{fairclip} includes 10,000 samples and FairVision\cite{fairvision} includes 30,000 subjects. These collections integrate image data with textual corpora and support interdisciplinary developments at the interface of computer vision and natural language processing for ophthalmic research.

\subsection*{Medical Imaging for Glaucoma}

In glaucoma research, medical imaging plays a central role in developing and validating automated diagnostic methods\cite{fairseg,fairvision,gdp,lag-1,cnn-1}. Even when sufficient data are available, challenges remain in labeling quality and consistency, since subjective variability among clinicians may introduce biases and limit model performance\cite{AIROGS}. Generative artificial intelligence (G-AI) provides a promising strategy to address data deficiencies and labeling inconsistencies, although human verification is still necessary and synthetic data may encounter similar annotation challenges\cite{gans,synthetic,review,impact,effects,cyclegan}. Among generative approaches, generative adversarial networks\cite{gan,gans,cyclegan,SegAN} have been particularly influential because of their ability to mitigate both data scarcity and labeling difficulties in an unsupervised manner. In practice, GAN-based segmentation pipelines are typically coupled with dedicated segmentation networks. For this segmentor component, pre-trained architectures such as convolutional neural networks\cite{lag-1,lag-2,cnn-1}, U-Net\cite{unet} and its variants\cite{unetpp,attunet,attention}, vision transformers\cite{vit,convit}, Mask R-CNN\cite{maskrcnn}, and more recently medical foundation models like MedSAM\cite{MedSAM} are frequently adopted to produce the final segmentation output. Recent work across both computer vision and medical imaging communities has consistently demonstrated the effectiveness of this paradigm for glaucoma-related imaging tasks.

\section*{Methodology}

\begin{figure}[ht]
  \centering
   \centerline{\includegraphics[width=0.8\columnwidth]{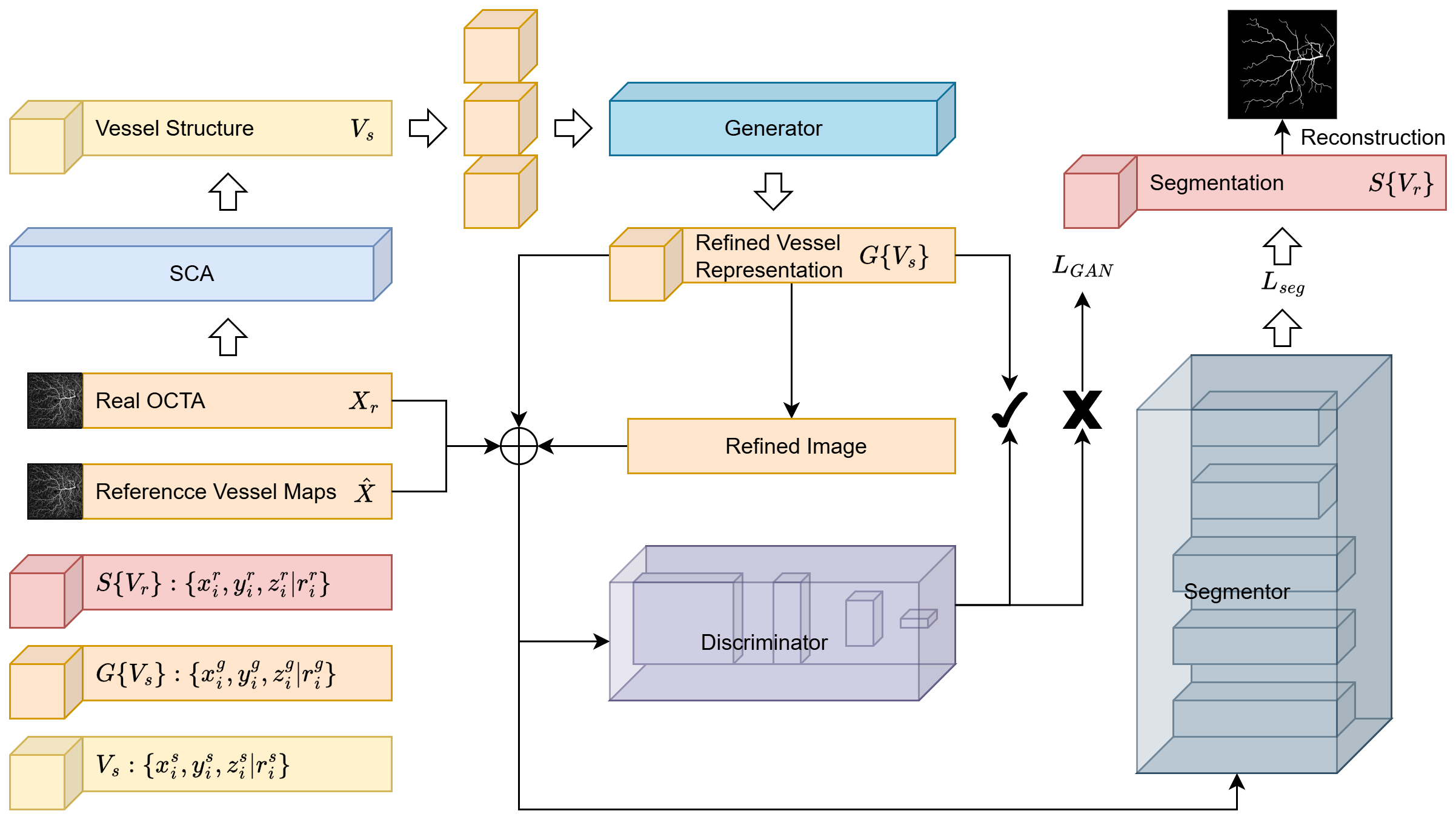}}
    \caption{The Architecture of BioVessel-Net}
    \label{s1}
\end{figure}

The complete framework of BioVessel-Net is illustrated in Figure~\ref{s1} where red indicates that BioVessel-Net does not utilize these things in Segmentor $S$, whereas G-AI+SLSMS do. and it consists of two parts: \textit{Vessel Structure Refinement Module} and \textit{Segmentor}.

\subsection*{Vessel Structure Refinement Module}

In OCTA imaging, retinal blood vessels exhibit a uniform color change and have a shape closely resembling a path. They can be represented using coordinates along with road width. Space Colonization Algorithm (SCA)\cite{sca} just meets this requirement. Rather than generating pixel-wise vessel segmentations, we formulates the vascular network as a structured graph representation, ensuring topological consistency and robust connectivity. The graph consists of the following components, shown in Equation~\ref{sca}:
\begin{equation}
  V = \left\{ (x_i, y_i, z_i | r_i) \right\}_{i=1}^{N}
  \label{sca}
\end{equation}
where ($x_{i}$, $y_{i}$, $z_i$) are the spatial coordinates of the vessel centerline points, $r_{i}$ represents the local vessel radius and $N$ means the number of nodes. Given a vessel structure $V_{s}$ via SCA, our generator $G$ learns to transform it into a realistic OCTA-like representation $V_r'$, expressed as $G(V_s)$, while $V_r'$ undergoes style adaptation to align with the contrast and noise characteristics of real OCTA images $V_r$. To enhance structural consistency, we modify the original CycleGAN architecture by removing the inverse generator and incorporating a segmentation consistency loss via the segmentor $S$. This modification prevents CycleGAN from altering the vessel topology, ensuring that vessel segmentation remains biologically meaningful, as illustrated in Equation~\ref{lu}:
\begin{equation}
    L_{\text{seg}} = \mathbb{E}_{V_r} \left[ \| S(V_r') - S(V_r) \|_1 \right]
   \label{lu}
\end{equation}
where $S(V)$ extracts vessel masks from OCTA images, enforcing structural consistency during transformation. This constraint ensures that CycleGAN adapts contrast while strictly preserving vessel topology. Consequently, our total loss function is formulated as follows:

\begin{align}
    L_{\text{GAN}} &= \mathbb{E}_{V_r} [D(V_r)] - \mathbb{E}_{V_s} [D(G(V_s))] 
    \label{eq:gan_loss} \\
    L_{\text{GP}} &= \lambda \mathbb{E}_{\hat{V}} \left[ \left( \|\nabla_{\hat{V}} D(\hat{V})\|_2 - 1 \right)^2 \right]
    \label{eq:gp_loss} \\
    L_{\text{total}} &= L_{\text{GAN}} + \lambda_{\text{GP}} L_{\text{GP}} + \lambda_{\text{seg}} L_{\text{seg}}
    \label{eq:total_loss}
\end{align}
where Equation~\ref{eq:gan_loss} is the Wasserstein Adversarial Loss, Equation~\ref{eq:gp_loss} is the Gradient Penalty for Stability ($\hat{X}$ is the reference vessel representation) and Equation~\ref{eq:total_loss} is the Structural Consistency via Segmentation Loss. Upon training convergence, the generator produces a vessel structure that not only exhibits realistic contrast properties but also preserves anatomical coherence, ensuring fidelity to real OCTA characteristics.

\subsection*{Segmentor}

Rather than utilizing pixel-wise segmentation maps, we directly extract primary vessels from the $(x_{i}^{G}, y_{i}^{G},z_{i}^{G}|r_{i}^{G})$ representation using DFS based graph traversal\cite{DFS}, which effectively isolates large vessels while filtering out capillaries.

First, we construct a graph $G=(V,E)$, where nodes $V$ represent vessel centerline points and edges $E$ connect adjacent vessel points based on local vessel connectivity. A vessel segment $e_{ij}$ between nodes $v_i$ and $v_j$ (the direction is from $i$ to $j$) is assigned a weight, as shown in Equation~\ref{we}:
\begin{equation}
    w(e_{ij}) = r_i
   \label{we}
\end{equation}
where larger vessels are prioritized in the traversal process. To extract primary vessels, we apply radius thresholding and DFS traversal\cite{DFS}. For radius filtering, we define a minimum main vessel radius $R_{min}$ and retain only Equation~\ref{rmin}:
\begin{equation}
    V_{\text{main}} = \left\{ (x_i, y_i, z_i | r_i) \mid r_i \geq (R_{\min}*r_{max}) \right\}
   \label{rmin}
\end{equation}

For DFS traversal, it consists of three steps:
\begin{itemize}
    \item select the optic disc region as the root node
    \item recursively traverse the largest connected component using DFS
    \item stop when no further vessel segments satisfy the radius constraint
\end{itemize}
when all three steps finished, construct final extracted vessel structure $G_{main}$. This approach effectively retains only the primary vascular structure, removing small capillaries and noise. This method can be summarized as the following Algorithm~\ref{alg:dfs_vessel_extraction}.

\begin{algorithm}[ht]
\caption{Primary Vessel Extraction via DFS}
\label{alg:dfs_vessel_extraction}
\begin{algorithmic}[1]
\REQUIRE Vessel Structure $V = \{(x_i, y_i, z_i \mid r_i)\}_{i=1}^{N}$
\ENSURE Extracted Primary Vessel Graph $G_{\text{main}}$

\STATE \textbf{Step 1: Vessel Graph Construction}
\FORALL{vessel point $(x_i, y_i, z_i \mid r_i) \in V$}
    \STATE Identify Neighboring Vessel Points to Form Edge Set $E$
    \STATE Assign Edge Weight $w(e_{ij}) = r_i$
\ENDFOR

\STATE \textbf{Step 2: Vessel Filtering}
\STATE Apply Radius Threshold $R_{\text{min}}$
\STATE $V_{\text{main}} = \{ (x_i, y_i, z_i \mid r_i) \mid r_i \geq R_{\text{min}} \cdot r_{\max} \}$
\STATE Remove Weakly Connected Components

\STATE \textbf{Step 3: DFS-Based Traversal}
\STATE Initialize DFS from Optic Disc Region
\FORALL{point $(x_i, y_i, z_i \mid r_i) \in V_{\text{main}}$}
    \STATE Recursively Explore Vessel Network via DFS
    \STATE Preferentially Expand along Thicker Vessels
\ENDFOR

\STATE \textbf{Step 4: Output}
\RETURN $G_{\text{main}}$

\end{algorithmic}
\end{algorithm}

\begin{figure}
    \centering
    \includegraphics[width=0.8\linewidth]{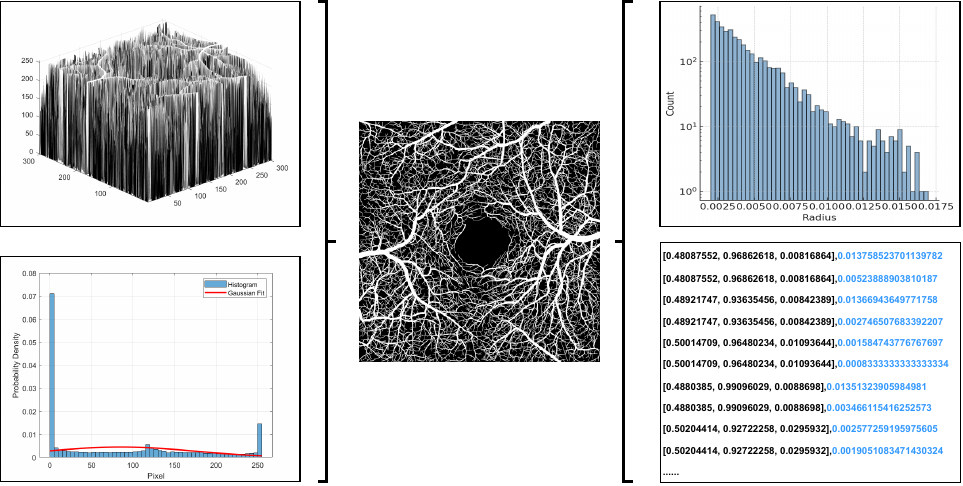}
    \caption{Details of One Sample: 3D Visualization, Statistical Distribution, Radius Statistics and Vascular Data}
    \label{me-2}
\end{figure}

As shown in Figure~\ref{me-2}, the key aspect of Algorithm~\ref{alg:dfs_vessel_extraction} is the threshold $R_{min}$. Instead of only examining the final image, we analyze intermediate outputs, CSV-derived statistics, and GAN-generated vessel radius distributions. For a representative image, we computed a Mean of 85.31 and a Standard Deviation of 88.67, which determined $R_{min}$ as 0.1968 $r_{max}$. Across the datasets, the corresponding values were 0.2011 (RetinaMix), 0.2102 (OCTA-500), and 0.1982 (ROSE). For consistency, $R_{min}$ was standardized to 0.2. Independent ophthalmology analyses confirm that OCTA images typically exhibit 18-26\% capillary coverage, supporting the validity of this threshold.

\section*{Dataset \& Implementation}

\subsection*{Dataset}

\subsubsection*{OCTA-500}
The OCTA-500\cite{octa500} is a large-scale retinal OCTA resource comprising 500 en face images acquired under two scanning protocols: a wide-field 6 mm $\times$ 6 mm view (300 images, ID 10001-10300) for global vascular assessment and a high-resolution 3 mm $\times$ 3 mm view (200 images, ID 10301-10500) for detailed fovea-centered microvascular analysis. Each image is accompanied by expert annotations that delineate arteries, veins, large vessels, capillaries, and both two-dimensional and three-dimensional foveal avascular zones (FAZ), providing comprehensive ground truth for vascular morphology studies. With its dual-scale design, OCTA-500 supports both wide-field examination of retinal circulation and fine-grained investigation of capillary networks, offering a standardized benchmark for vessel segmentation, classification, and FAZ analysis in the study of retinal structure and disease progression.

\subsubsection*{ROSE}

The ROSE\cite{ROSE} is the first publicly available benchmark dedicated to vessel segmentation in OCTA images. It comprises 229 images with expert manual annotations at either pixel or centerline level. The dataset includes two subsets: ROSE-1, containing 117 images from 39 subjects (26 with Alzheimer’s disease and 13 healthy controls) with both pixel-level and centerline annotations; and ROSE-2, consisting of 112 images from 112 eyes with macular diseases, annotated at the centerline level. The images were captured using two different OCTA systems (Optovue RTVue XR Avanti and Heidelberg OCT2), both covering a 3 $\times$ 3 mm$^2$ fovea-centered region. ROSE provides a standardized resource for training and evaluating OCTA vessel segmentation methods and has been applied to investigate retinal microvascular biomarkers in neurodegenerative diseases.

\subsubsection*{RetinaMix}

\begin{figure}[ht]
  \centering
  \begin{subfigure}{0.24\linewidth}
    \centerline{\includegraphics[width=0.8\columnwidth]{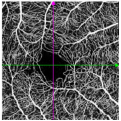}}
    \caption{Intalight}
    \label{i1-Angio}
  \end{subfigure}
  \hfill
  \begin{subfigure}{0.24\linewidth}
    \centerline{\includegraphics[width=0.8\columnwidth]{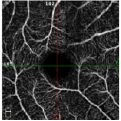}}
    \caption{OPTOVUE}
    \label{o1-Angio}
  \end{subfigure}
  \hfill
  \begin{subfigure}{0.24\linewidth}
    \centerline{\includegraphics[width=0.8\columnwidth]{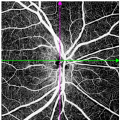}}
    \caption{Intalight}
    \label{i1-s}
  \end{subfigure}
    \hfill
  \begin{subfigure}{0.24\linewidth}
    \centerline{\includegraphics[width=0.8\columnwidth]{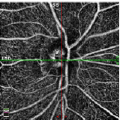}}
    \caption{OPTOVUE}
    \label{o1-s}
  \end{subfigure}
  \caption{Comparison of Data Collected by Different Devices for the Same Patient } 
  \label{dc}
\end{figure}

RetinaMix is a unique and pioneering dataset focused exclusively on retinal vascularity. The dataset features an exceptionally clear distribution of capillaries. It comprises 550 images, both 2D and 3D, from 250 subjects, ensuring fair representation across gender and race. We collected retinal vascular data for both eyes (OS and OD) of each subject, with a gender split of 48.8\% male and 51.2\% female. Racial distribution reflects local glaucoma clinic demographics: 80.4\% white, 14.6\% Black, with Asians and other minority groups comprising the remainder. Additional data includes age (47.1 $\pm$ 24.8 years) and examination dates from 2023 to 2024. RetinaMix's original images come from the latest Intalight device\footnote{\url{https://intalight.com/}}, offering superior detail compared to OPTOVUE\footnote{\url{https://www.visionix.com/}} (Figure~\ref{dc}). As shown in Figure~\ref{i1-Angio} and Figure~\ref{i1-s}, Intalight images feature clearer vessel boundaries and capillary distributions, enhancing segmentation accuracy. In supervised learning, their minimal pixel variation introduces valuable noise for testing model robustness. For 3D reconstruction and evolution prediction, these high-resolution images aid in vascular modeling and blood flow analysis, making Intalight data ideal for both tasks.

\begin{figure}[ht]
  \centering
      \begin{subfigure}{0.3\linewidth}
    \centerline{\includegraphics[width=\columnwidth]{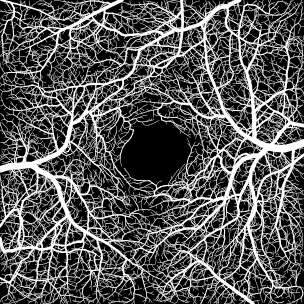}}
    \caption{RetinaMix}
    \label{cid-a}
  \end{subfigure}
    \hfill
  \begin{subfigure}{0.3\linewidth}
\centerline{\includegraphics[width=\columnwidth]{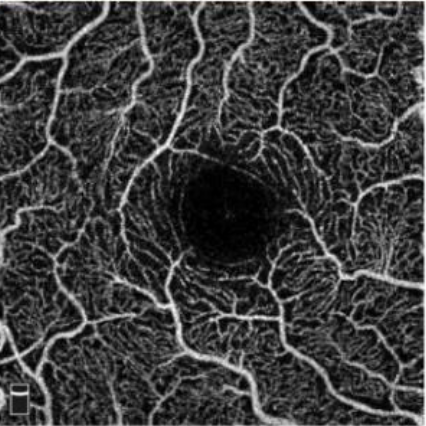}}
    \caption{OCTA-500}
    \label{cid-c}
  \end{subfigure}
  \hfill
  \begin{subfigure}{0.3\linewidth}
\centerline{\includegraphics[width=\columnwidth]{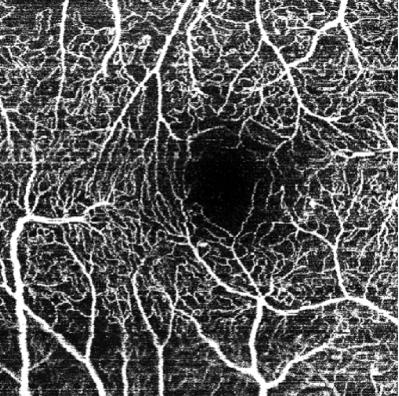}}
    \caption{ROSE}
    \label{cid-b}
  \end{subfigure}
  \caption{Comparison of RetinaMix with Other Datasets}
  \label{comparison_image_dataset}
\end{figure}

For RetinaMix, we eliminate red and green baselines from Intalight device images through a multi-step process. First, we apply a dark channel algorithm \cite{dcp} to remove light artifacts, followed by unsharp masking \cite{usm} to enhance high-frequency details. The final images (Figure~\ref{cid-a}) undergo manual verification for quality assurance. As shown in Figure~\ref{comparison_image_dataset}, while main vessels remain visible across datasets, magnification reveals that OCTA-500 \cite{octa500} and ROSE \cite{ROSE} blur capillaries and vessel ends. Even the unprocessed baseline image (Figure~\ref{o1-Angio}) offers superior vessel clarity compared to other datasets (Figure~\ref{cid-b}, Figure~\ref{cid-c}). Among them, Figure~\ref{i1-Angio} and Figure~\ref{o1-Angio} are images of deep angio, and Figure~\ref{i1-s} and Figure~\ref{o1-s} are images of superficial angio. They all are images of inner retina vessel.

\begin{figure}[ht]
  \centering
  \begin{subfigure}{0.32\linewidth}
    \centerline{\includegraphics[width=0.8\columnwidth]{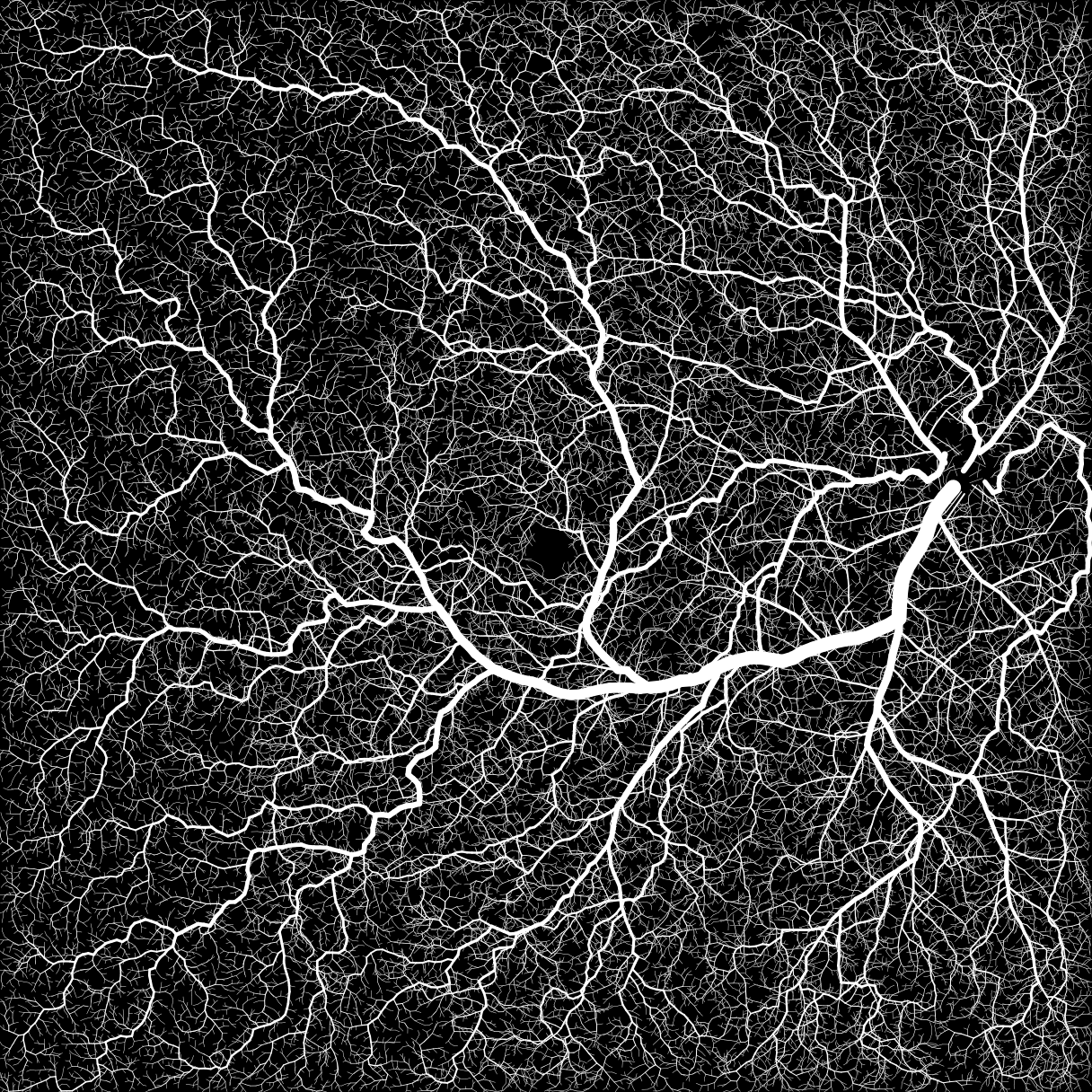}}
    \caption{Top View}
    \label{2d}
  \end{subfigure}
    \hfill
    \begin{subfigure}{0.32\linewidth}
\centerline{\includegraphics[width=0.8\columnwidth]{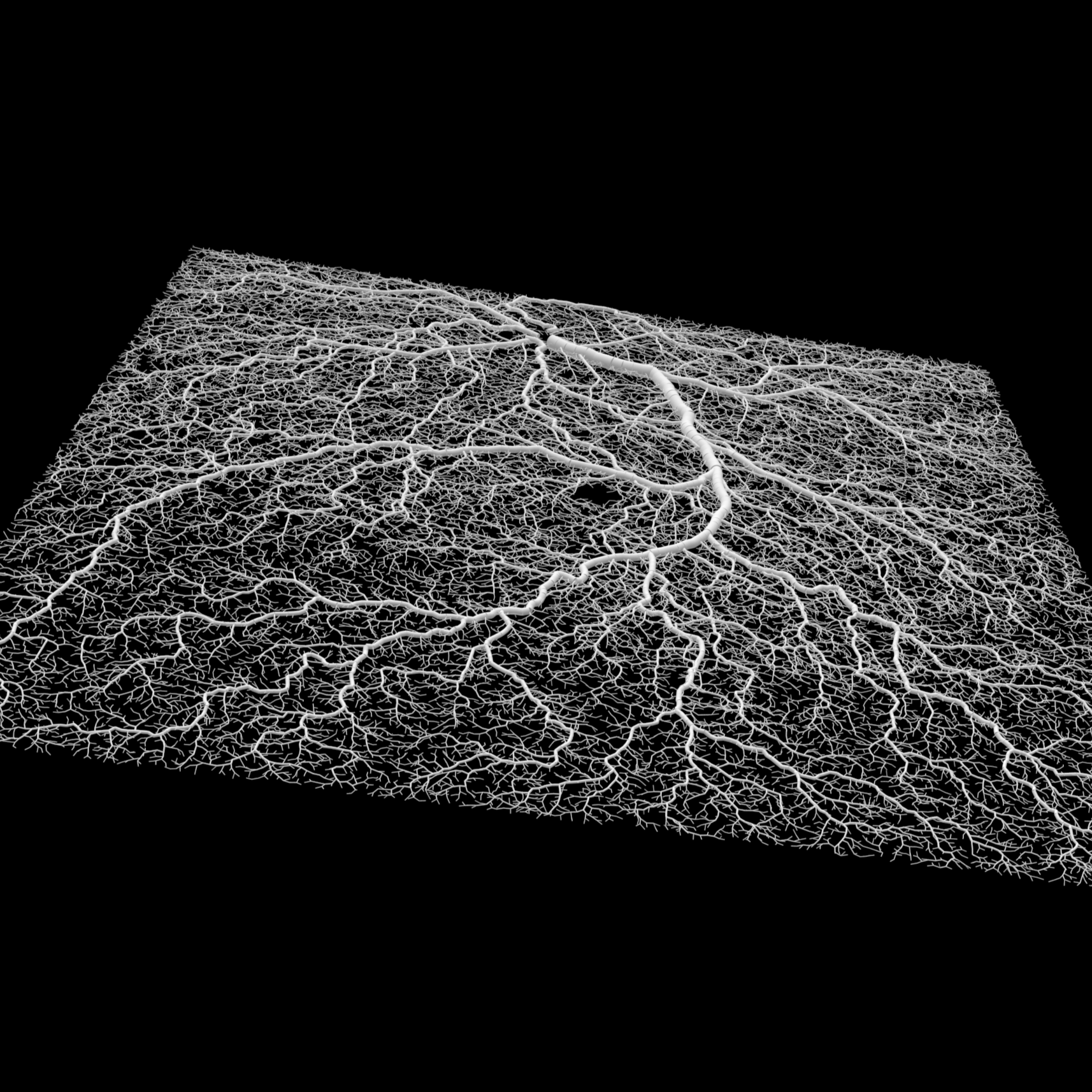}}
    \caption{Front View}
    \label{3d-2}
  \end{subfigure}
    \hfill
      \begin{subfigure}{0.32\linewidth}
    \centerline{\includegraphics[width=0.8\columnwidth]{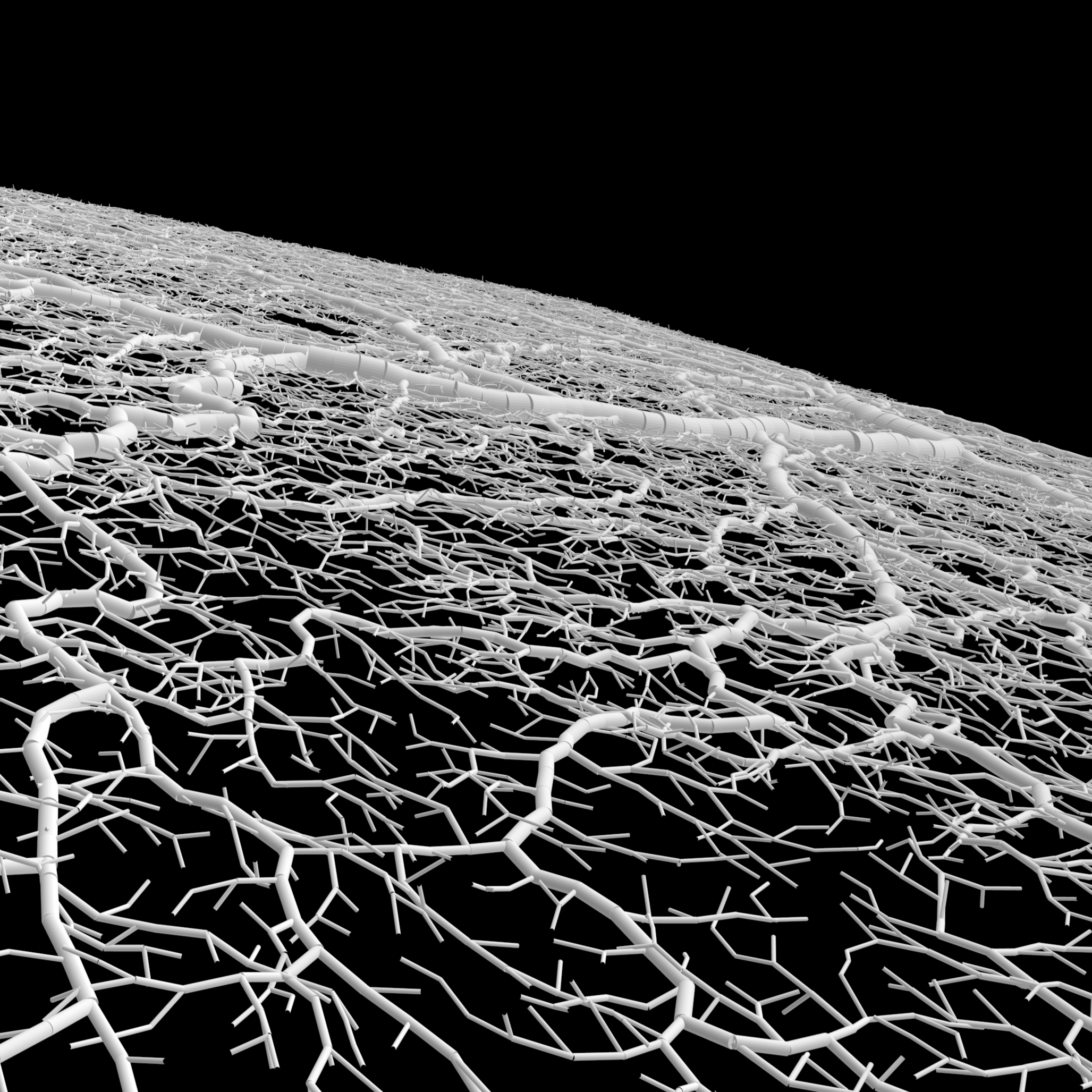}}
    \caption{Detail}
    \label{3d}
  \end{subfigure}
  \caption{3D Imaging of Retinal Vessel from Various Angles}
  \label{3D}
\end{figure}

For the representation of 3D imaging of RetinaMix, as shown in Figure~\ref{3D}, capillaries stand out from the thicker main blood vessels by attaching to them and narrowing at their ends (Figure~\ref{3d}). The elaborate architecture of blood vessels and the intertwining capillaries contribute to an extremely intricate vascular layout. 3D imaging part of RetinaMix is formatted as a JSON file, which includes 3D X, Y, and Z coordinates and the corresponding radius pixel distribution values. For annotation (2D, image format: PNG), our team includes two glaucoma specialists who handle labeling, verification, and manual evaluation, utilizing the Labelme\footnote{\url{https://github.com/wkentaro/labelme}}.

In general, RetinaMix is a dataset that combines 2D and 3D images with extremely high image resolution, and is particularly dedicated to retinal vascular segmentation.

\subsection*{Implementation}

\subsubsection*{Hyperparameter}

For BioVessel-Net, its generator adopts a ResNet 9-block architecture\cite{resnet}, while the discriminator utilizes PatchGAN\cite{pix2pix} (70$\times$70). The optimizer is Adam\cite{adam} with a learning rate of $2\times10^{-4}$ and momentum parameters $\beta_1=0.5$, $\beta_2=0.999$. BioVessel-Net is trained for 50 epochs, with a linear learning rate decay to 0 after 25 epochs, with evaluation performed using 10-fold cross-validation. Data augmentation includes: random rotations ($k\times90^\circ\pm10^\circ$), flipping and contrast adjustments.

\subsubsection*{Evaluation Metric}

We use \textit{Intersection over Union}\cite{IoU} (IoU), also known as the \textit{Jaccard Index}, to measure similarity between the predicted segmentation and the ground truth. This metric ranges from 0 to 1, with 1 indicating perfect segmentation. Additionally, we use the \textit{Dice Coefficient}\cite{Dice}, another metric for segmentation accuracy that calculates overlap between the prediction and ground truth. All in all, IoU is used to evaluate segmentation accuracy of the main vessels, while the Dice Coefficient assesses accuracy at finer vessel ends relative to the main vessels. For vascular structural enhancement, we use the \textit{Structural Similarity Index}\cite{SSIM} (SSIM) and the \textit{Mean Squared Error}\cite{MSE} (MSE). 
SSIM (range: $-1$ to $1$) assesses similarity based on luminance, contrast, and structure, while MSE quantifies pixel-wise differences, with lower values indicating higher similarity. While SSIM aligns with human perception, MSE focuses on absolute pixel differences. 

Finally, the best results in each column are highlighted in bold, and the second-best results are indicated with underline.

\subsubsection*{Hardware}
The hardware specifications for training and testing include 2 Tesla V100 GPUs (2 $\times$ 32GB), 64GB of RAM, 8 CPU cores per node, and a total of 6 nodes.

\section*{Experimental Result}

\subsection*{Baseline Model}
All final experimental results are the average of ten cross-validation times. All selected models' hyperparameters have been optimized by Optuna\footnote{\url{https://optuna.org/}}. Finally, we selected many SOTA models for experimental verification:

\subsubsection*{CNN-based Model}
We included several classic convolutional neural network (CNN) architectures: U-Net\cite{unet}, U-Net++\cite{unetpp}, SegNet\cite{SegNet}, DeepLabV3+\cite{DeepLabV3+}, Mask R-CNN\cite{maskrcnn}, and YOLOv11-x\cite{v11}. These models represent widely adopted backbones for medical image segmentation tasks.

\subsubsection*{Transformer-based Model}
To incorporate recent advances in transformer architectures, we evaluated TransUNet\cite{TransUNet}, Attention U-Net\cite{attunet} and Swin-UNet\cite{Swin-UNet}, which combine self-attention mechanisms with encoder-decoder designs.

\subsubsection*{Large-scale Pretrained Model}
We further benchmarked foundation models designed for general-purpose segmentation, including SAM\cite{SAM} and its medical adaptation MedSAM\cite{MedSAM}.

\subsubsection*{Self-supervised and Universal Framework}
Representative frameworks in this category include CauSSL\cite{CauSSL}, UniverSeg\cite{UniverSeg}, S2VNet\cite{S2VNet}, and Tyche\cite{tyche}, which leverage self-supervised or universal segmentation strategies to improve generalizability across domains.

\subsubsection*{G-AI based Method}
We incorporated adversarial segmentation models such as SegAN\cite{SegAN}, Vessel-GAN\cite{Vessel-GAN}, LGGAN\cite{LGGAN}, CSGAN\cite{CSGAN}, CycleGAN-Seg\cite{CycleGAN-Seg}, CycleGAN\cite{cyclegan}, and a generic 
GAN baseline\cite{gan}. These approaches use adversarial training to enhance boundary accuracy and robustness.

\subsubsection*{Ophthalmology-specific Model}
Finally, we included OCTA-Net\cite{OCTA-Net}, which is specifically designed for retinal and choroidal vessel segmentation in OCTA images.

\subsection*{Comparative Experiment}
\subsubsection*{Benchmarking BioVessel-Net Against Competing Approaches}

As shown in Table \ref{er-1}, across all datasets, BioVessel-Net consistently achieves the best performance, with IoU and Dice scores surpassing 99\%. Particularly, on RetinaMix (6 mm), it records 99.41 IoU and 99.71 Dice, and on OCTA-500 (3 mm) it achieves 99.42 IoU and 99.71 Dice, markedly outperforming the strongest baselines such as MedSAM\cite{MedSAM} and OCTA-Net\cite{OCTA-Net}. The improvement is more pronounced in cross-dataset generalization, where BioVessel-Net demonstrates superior robustness compared with both CNN-\cite{unet,unetpp,SegNet,DeepLabV3+,maskrcnn,v11} and transformer-based methods\cite{Swin-UNet,attunet,TransUNet}. These results highlight the effectiveness of incorporating biostatistical priors into a GAN-driven segmentation framework, yielding both high accuracy and strong generalizability.
\begin{table*}[ht]
\begin{center}
\caption{Comparison BioVessel-Net With Other Models on RetinaMix, OCTA-500 and ROSE ($\times$100\%)}
\begin{tabular}{l|cc|cc|cc|cc|cc}
\hline
\multirow{4}{*}{\textbf{Model}}  & \multicolumn{10}{c}{\textbf{Dataset}} \\
\cline{2-11}  &  \multicolumn{4}{c|}{\textbf{RetinaMix}} & \multicolumn{4}{c|}{\textbf{OCTA-500}} & \multicolumn{2}{c}{\textbf{ROSE}}\\
\cline{2-11}  & \multicolumn{2}{c|}{\textbf{6mm}} & \multicolumn{2}{c|}{\textbf{3mm}} & \multicolumn{2}{c|}{\textbf{6mm}} & \multicolumn{2}{c|}{\textbf{3mm}} & \multicolumn{2}{c}{\textbf{3mm}} \\

\cline{2-11}  &  \textbf{IoU} & \textbf{Dice} & \textbf{IoU} & \textbf{Dice} & \textbf{IoU} & \textbf{Dice} & \textbf{IoU} & \textbf{Dice} & \textbf{IoU} & \textbf{Dice} \\
\hline
U-Net\cite{unet}                  & 95.14 & 97.51 & 95.23 & 97.56 & 96.93 & 98.44 & 98.75 & 99.37 & \underline{97.94} & \underline{98.96} \\ 
U-Net++\cite{unetpp}              & 95.68 & 97.81 & 96.64 & 98.29 & \underline{97.65} & \underline{98.81} & 97.96 & 98.97 & 97.52 & 98.75 \\ 
Attention U-Net\cite{attunet}       & 95.78 & 97.84 & 96.04 & 97.98 & 97.05 & 98.50 & 98.61 & 99.30 & 97.92 & 98.95 \\
Mask R-CNN\cite{maskrcnn}             & 92.35 & 96.02 & 91.35 & 95.48 & 92.06 & 95.87 & 93.50 & 96.64 & 91.23 & 95.41 \\
YOLOv11-x\cite{v11}               & 95.81 & 97.86 & \underline{96.55} & \underline{98.24} & 95.93 & 97.92 & 97.66 & 98.82 & 97.83 & 98.89 \\ 
MedSAM\cite{MedSAM}               & 96.38 & 98.16 & 96.47 & 98.20 & 96.01 & 97.95 & \underline{98.94} & \underline{99.47} & 97.89 & 98.93 \\ 
CauSSL\cite{CauSSL}               & 95.23 & 97.56 & 95.47 & 97.68 & 96.32 & 98.04 & 98.15 & 99.08 & 97.80 & 98.89 \\ 
UniverSeg\cite{UniverSeg}         & 95.30 & 97.58 & 95.52 & 97.70 & 96.58 & 98.17 & 98.42 & 99.22 & 97.07 & 98.01 \\
S2VNet\cite{S2VNet}               & 95.68 & 97.81 & 95.36 & 97.63 & 96.45 & 98.11 & 98.27 & 99.15 & 97.92 & 98.95 \\
Tyche\cite{tyche}                 & 95.80 & 97.85 & 95.19 & 97.53 & 96.28 & 98.02 & 98.11 & 99.06 & 97.76 & 98.87 \\ 
DeepLabV3+\cite{DeepLabV3+} & 95.72 & 97.68 & 95.38 & 97.62 & 96.21 & 98.05 & 97.82 & 98.95 & 97.55 & 98.81 \\
SegNet\cite{SegNet} & 95.30 & 97.41 & 95.02 & 97.25 & 95.84 & 97.96 & 97.40 & 98.72 & 97.21 & 98.56 \\
TransUNet\cite{TransUNet} & 96.42 & 98.09 & 96.13 & 98.00 & 97.11 & 98.44 & 98.23 & 99.07 & 97.86 & 98.92 \\
Swin-UNet\cite{Swin-UNet} & 96.51 & 98.15 & 96.24 & 98.05 & 97.19 & 98.50 & 98.27 & 99.11 & 97.91 & 98.94 \\
OCTA-Net\cite{OCTA-Net} & \underline{96.68} & \underline{98.23} & 96.39 & 98.12 & 97.35 & 98.58 & 98.32 & 99.13 & \underline{97.94} & \underline{98.96} \\
SAM\cite{SAM} & 96.22 & 98.01 & 95.95 & 97.86 & 97.02 & 98.36 & 98.16 & 99.03 & 97.82 & 98.88 \\
SegAN\cite{SegAN} & 95.61 & 97.55 & 95.32 & 97.43 & 96.08 & 98.02 & 97.58 & 98.81 & 97.36 & 98.62 \\
Vessel-GAN\cite{Vessel-GAN} & 96.37 & 98.07 & 96.08 & 97.94 & 97.14 & 98.42 & 98.20 & 99.06 & 97.89 & 98.91 \\
LGGAN\cite{LGGAN} & 95.74 & 97.66 & 95.45 & 97.53 & 96.25 & 98.09 & 97.69 & 98.87 & 97.48 & 98.71 \\
CSGAN\cite{CSGAN} & 95.80 & 97.70 & 95.51 & 97.58 & 96.33 & 98.12 & 97.73 & 98.90 & 97.52 & 98.74 \\
CycleGAN-Seg\cite{CycleGAN-Seg} & 95.91 & 97.75 & 95.63 & 97.64 & 96.41 & 98.18 & 97.79 & 98.92 & 97.59 & 98.77 \\

\hline
BioVessel-Net                             & \textbf{99.41} & \textbf{99.71} & \textbf{98.66} & \textbf{99.33} & \textbf{99.21} & \textbf{99.60} & \textbf{99.42} & \textbf{99.71} & \textbf{99.19} & \textbf{99.59} \\ 
\hline
\end{tabular}
\label{er-1}
\end{center}
\end{table*}

\subsubsection*{Evaluating Cross-Dataset Robustness of BioVessel-Net}

Table~\ref{er-2-1} reports mixed validation results. When trained on RetinaMix and tested on OCTA-500, BioVessel-Net achieved 99.21 IoU and 99.60 Dice, higher than MedSAM\cite{MedSAM} (97.95/98.84), OCTA-Net\cite{OCTA-Net} (98.36/98.92), and Swin-UNet\cite{Swin-UNet} (98.10/98.71). In the reverse setting (OCTA-500 → RetinaMix), BioVessel-Net still reached 99.41/99.70, outperforming U-Net++\cite{unetpp} (98.78/99.40), TransUNet\cite{TransUNet} (98.63/99.17), and Tyche\cite{tyche} (98.12/99.08). On ROSE, BioVessel-Net kept 99.55/99.78, clearly ahead of other methods such as YOLOv11-x\cite{v11} (98.72/99.42) and DeepLabV3+\cite{DeepLabV3+} (98.43/99.21). These results confirm that BioVessel-Net is consistently stronger than CNN baselines (U-Net\cite{unet}, SegNet\cite{SegNet}), transformer models (TransUNet\cite{TransUNet}, Swin-UNet\cite{Swin-UNet}), and GAN-based methods (SegAN\cite{SegAN}, Vessel-GAN\cite{Vessel-GAN}, CycleGAN-Seg\cite{CycleGAN-Seg}).

\begin{table*}[ht]
\centering
\caption{Conducting Mixed Validation on Models using Mixed Datasets ($\times$100\%) }
\scalebox{0.75}{
\begin{tabular}{l|cc|cc|cc|cc|cc|cc|cc|cc}
\hline
\multirow{5}{*}{\textbf{Model}}  & \multicolumn{16}{c}{\textbf{Dataset}} \\
\cline{2-17}  &  \multicolumn{4}{c|}{\textbf{ROSE}} &  \multicolumn{2}{c|}{\textbf{RetinaMix}} & \multicolumn{2}{c|}{\textbf{OCTA-500}} &  \multicolumn{4}{c|}{\textbf{RetinaMix} } & \multicolumn{4}{c}{\textbf{OCTA-500} }\\
\cline{2-17}  & \multicolumn{2}{c|}{\textbf{RetinaMix}} & \multicolumn{2}{c|}{\textbf{OCTA-500}} & \multicolumn{4}{c|}{\textbf{ROSE}} &  \multicolumn{4}{c|}{\textbf{OCTA-500} } & \multicolumn{4}{c}{\textbf{RetinaMix} }  \\
\cline{2-17} &  \multicolumn{8}{c|}{\textbf{3mm}} & \multicolumn{2}{c|}{\textbf{6mm}} & \multicolumn{2}{c|}{\textbf{3mm}} & \multicolumn{2}{c|}{\textbf{6mm}} & \multicolumn{2}{c}{\textbf{3mm}}\\
\cline{2-17}  &  \textbf{IoU} & \textbf{Dice} & \textbf{IoU} & \textbf{Dice} &  \textbf{IoU} & \textbf{Dice} & \textbf{IoU} & \textbf{Dice} &  \textbf{IoU} & \textbf{Dice} & \textbf{IoU} & \textbf{Dice} & \textbf{IoU} & \textbf{Dice} & \textbf{IoU} & \textbf{Dice} \\
\hline
U-Net\cite{unet}                & 98.34 & 99.16 & 98.32 & 99.15 & 93.89 & 96.85 & 94.23 & 97.03 & 96.21 & 98.07 & 95.23 & 97.56 & 96.98 & 98.47 & 96.81 & 98.38 \\
U-Net++\cite{unetpp}            & 98.99 & 99.49 & 98.78 & 99.39 & 94.08 & 96.95 & 94.46 & 97.15 & 95.73 & 97.82 & 97.01 & 98.51 & 97.59 & 98.78 & 96.93 & 98.44\\
Attention U-Net\cite{attention}     & 98.54 & 99.26 & 98.23 & 99.11 & 94.07 & 96.94 & 94.52 & 97.18& 96.99 & 98.47 & 96.09 & 98.01 & 97.21 & 98.59 & 96.78 & 98.36\\
Mask R-CNN\cite{maskrcnn}           & 96.37 & 98.15 & 96.31 & 98.12 & 90.73 & 95.14  & 91.89 & 95.77& 94.23 & 97.03 & 94.11 & 96.97 & 96.68 & 98.31 & 95.96 & 97.94\\
YOLOv11-x\cite{v11}            & 98.88 & 99.44 & 98.72 & 99.36 & 94.11 & 96.97 & 94.49 & 97.17& 97.32 & 98.64 & 96.54 & 98.24 & 96.12 & 98.02 & 96.63 & 98.29\\
MedSAM\cite{MedSAM}             & 99.04 & 99.52 & \underline{98.98} & \underline{99.49} & 95.01 & 97.41 & 95.42 & 97.66 & 97.54 & 98.75 & 95.87 & 97.89 & \underline{98.23} & \underline{99.11} & 97.46 & 98.72 \\
CauSSL\cite{CauSSL}             & 96.83 & 98.38 & 95.89 & 97.90 & 96.98 & 98.47 & 97.83 & 98.88 & \textbf{99.12} & \textbf{99.56} & \underline{98.79} & \underline{99.39} & 94.87 & 97.33 & 95.34 & 97.62\\
UniverSeg\cite{UniverSeg}      & 98.96 & 99.48 & 98.56 & 99.28 & 94.23 & 97.03 & 94.59 & 97.24  & 96.81 & 98.38 & 96.32 & 98.16 & 97.91 & 98.96 & \underline{98.12} & \underline{99.05}\\
S2VNet\cite{S2VNet}             & \underline{99.23} & \underline{99.61} & 98.79 & 99.39 & 93.89 & 96.85 & 95.22 & 97.55 & 96.95 & 98.44 & 95.83 & 97.87 & 97.87 & 98.95 & 97.87 & 98.95\\
Tyche\cite{tyche}               & 98.78 & 99.38 & 98.43 & 99.21 & 94.22 & 97.02 & 94.98 & 97.46 & 96.88 & 98.41 & 96.11 & 98.02 & 97.84 & 98.94 & 97.68 & 98.87\\
DeepLabV3+\cite{DeepLabV3+} & 98.72 & 99.48 & 98.39 & 99.40 & 96.92 & 98.31 & 97.46 & 98.91 & 97.05 & 98.47 & 96.89 & 98.35 & 97.62 & 98.99 & 96.93 & 98.28 \\
SegNet\cite{SegNet}     & 98.21 & 99.12 & 97.96 & 99.00 & 96.41 & 98.07 & 97.03 & 98.64 & 96.58 & 98.18 & 96.23 & 97.95 & 97.11 & 98.72 & 96.36 & 97.89 \\
TransUNet\cite{TransUNet}  & 98.95 & 99.57 & 98.66 & 99.51 & 97.24 & 98.66 & 98.05 & 99.18 & 97.63 & 98.80 & 97.18 & 98.62 & 98.14 & 99.23 & 97.26 & 98.55 \\
Swin-UNet\cite{Swin-UNet}  & 98.98 & 99.59 & 98.70 & 99.53 & \underline{97.31} & \underline{98.71} & \underline{98.09} & \underline{99.22} & 97.69 & 98.84 & 97.25 & 98.67 & 98.18 & 99.26 & 97.32 & 98.59 \\
OCTA-Net\cite{OCTA-Net}   & 99.04 & 99.62 & 98.75 & 99.56 & 96.98 & 98.34 & 97.69 & 98.83 & 97.74 & 98.87 & 97.31 & 98.70 & 98.22 & 99.29 & 97.37 & 98.63 \\
SAM\cite{SAM}        & 98.84 & 99.51 & 98.55 & 99.45 & 97.10 & 98.58 & 97.92 & 99.10 & 97.51 & 98.73 & 97.05 & 98.54 & 97.96 & 99.14 & 97.12 & 98.49 \\
SegAN\cite{SegAN}      & 98.43 & 99.27 & 98.14 & 99.18 & 96.72 & 98.29 & 97.55 & 98.89 & 97.16 & 98.52 & 96.79 & 98.22 & 97.61 & 98.93 & 96.86 & 98.17 \\
Vessel-GAN\cite{Vessel-GAN} & 98.90 & 99.54 & 98.61 & 99.49 & 97.20 & 98.64 & 98.01 & 99.16 & 97.60 & 98.78 & 97.15 & 98.60 & 98.08 & 99.20 & 97.22 & 98.53 \\
LGGAN\cite{LGGAN}      & 98.58 & 99.36 & 98.29 & 99.28 & 96.88 & 98.40 & 97.72 & 98.97 & 97.30 & 98.63 & 96.94 & 98.36 & 97.79 & 99.02 & 97.02 & 98.31 \\
CSGAN\cite{CSGAN}      & 98.63 & 99.39 & 98.34 & 99.32 & 96.94 & 98.43 & 97.77 & 99.01 & 97.36 & 98.66 & 97.00 & 98.39 & 97.83 & 99.05 & 97.08 & 98.34 \\
CycleGAN-Seg\cite{CycleGAN-Seg} & 98.69 & 99.43 & 98.40 & 99.36 & 97.01 & 98.48 & 97.84 & 99.04 & 97.42 & 98.70 & 97.06 & 98.44 & 97.90 & 99.08 & 97.14 & 98.38 \\

\hline
BioVessel-Net                           & \textbf{99.55}& \textbf{99.78} & \textbf{99.21} & \textbf{99.60} & \textbf{97.38} & \textbf{98.76} & 
\textbf{98.14} & \textbf{99.25} & 
\underline{98.87} & \underline{99.43} & 
\textbf{98.81} & \textbf{99.40} & 
\textbf{99.58} & \textbf{99.79} & 
\textbf{99.41} & \textbf{99.70}\\
\hline
\multicolumn{17}{l}{\textbf{Note:} the second row corresponds to the model test set, while the third row represents the training set.}
\end{tabular}}
\label{er-2-1}
\end{table*}

\subsection*{Ablation Experiment}

\subsubsection*{Component-Wise Ablation of BioVessel-Net}

For baseline GAN model\cite{gan}, GAN\cite{gan} and CycleGAN\cite{cyclegan} are selected. For baseline segmentor, U-Net\cite{unet}, U-Net++\cite{unetpp}, Attention U-Net\cite{attunet} and MedSAM\cite{MedSAM} are selected. For BioVessel-Net, the dirrerent values of $R_{min}$ are tested to evaluate the model segmentation accuracy. As shown in Table~\ref{er-ab}, since main blood vessels make up a small proportion, the value calculated is also relatively low. When the minimum radius ratio is found, like BioVessel-Net, aligning with biostatistical properties, segmentation accuracy peaks, reinforcing the model’s strong clinical interpretability.

\begin{table*}[h]
\begin{center}
\caption{Ablation Experiment of BioVessel-Net ($\times$100\%)}
\begin{tabular}{ll|cc|cc|cc|cc|cc}
\hline
\multicolumn{2}{l|}{\multirow{4}{*}{\textbf{Model}}}  & \multicolumn{10}{c}{\textbf{Dataset}} \\
\cline{3-12} &   &\multicolumn{4}{c|}{\textbf{RetinaMix}} &\multicolumn{4}{c|}{\textbf{OCTA-500}} &\multicolumn{2}{c}{\textbf{ROSE}} \\
\cline{3-12} &   & \multicolumn{2}{c|}{\textbf{6mm}} & \multicolumn{2}{c|}{\textbf{3mm}} & \multicolumn{2}{c|}{\textbf{6mm}} & \multicolumn{2}{c|}{\textbf{3mm}} & \multicolumn{2}{c}{\textbf{3mm}}\\
\cline{3-12} &   &\textbf{IoU} & \textbf{Dice} & \textbf{IoU} & \textbf{Dice} &\textbf{IoU} & \textbf{Dice} & \textbf{IoU} & \textbf{Dice} & \textbf{IoU} & \textbf{Dice} \\
\hline
\multicolumn{1}{l|}{GAN\cite{gan}} & U-Net\cite{unet}             & 92.89 & 96.31 & 92.44 & 96.07 & 93.31 & 96.56 & 92.84 & 96.31 & 93.37 & 96.59 \\
\multicolumn{1}{l|}{GAN\cite{gan}} & U-Net++\cite{unetpp}           & 93.62 & 96.70 & 92.87 & 96.30 & 93.72 & 96.73 & 93.03 & 96.36 & 92.84 & 96.31 \\
\multicolumn{1}{l|}{GAN\cite{gan}} & Attention U-Net\cite{attunet}   & 93.56 & 96.67 & 93.12 & 96.44 & 93.65 & 96.70 & 92.85 & 96.31 & 93.42 & 96.61 \\
\multicolumn{1}{l|}{GAN\cite{gan}} & MedSAM\cite{MedSAM}            & 93.73 & 96.76 & 92.76 & 96.24 & 92.83 & 96.30 & 92.47 & 96.10 & 93.76 & 96.75 \\
\multicolumn{1}{l|}{GAN\cite{gan}} & DFS\cite{DFS}               & 95.81 & 97.86 & 95.24 & 97.56 & 94.83 & 97.33 & 94.87 & 97.35 & 94.73 & 97.27 \\
\hline
\multicolumn{1}{l|}{CycleGAN\cite{cyclegan}}  & U-Net\cite{unet}             & 93.44 & 96.61 & 93.42 & 96.60 & 94.73 & 97.26 & 93.97 & 96.89 & 93.76 & 96.75 \\
\multicolumn{1}{l|}{CycleGAN\cite{cyclegan}}  & U-Net++\cite{unetpp}           & 93.72 & 96.76 & 93.62 & 96.70 & 94.58 & 97.18 & 94.36 & 97.05 & 93.19 & 96.46 \\
\multicolumn{1}{l|}{CycleGAN\cite{cyclegan}}  & Attention U-Net\cite{attunet}   & 93.79 & 96.80 & 93.45 & 96.61 & 94.59 & 97.18 & 94.29 & 97.01 & 94.03 & 96.89 \\
\multicolumn{1}{l|}{CycleGAN\cite{cyclegan}}  & MedSAM\cite{MedSAM}            & 94.02 & 96.92 & 94.37 & 97.10 & 93.83 & 96.72 & 94.37 & 97.19 & 94.18 & 96.96 \\
\multicolumn{1}{l|}{CycleGAN\cite{cyclegan}}  & DFS\cite{DFS}               & 96.23 & 98.08 & 96.42 & 98.18 & 96.68 & 98.33 & 96.27 & 98.13 & 95.83 & 97.84 \\
\hline
\multicolumn{1}{l|}{CycleGAN$^{*}$}  & U-Net\cite{unet}             & 94.56 & 97.20 & 94.52 & 97.18 & 94.86 & 97.31 & 94.79 & 97.27 & 94.19 & 96.94 \\
\multicolumn{1}{l|}{CycleGAN$^{*}$}  & U-Net++\cite{unetpp}           & 94.98 & 97.43 & 94.36 & 97.10 & 94.83 & 97.29 & 95.09 & 97.41 & 94.27 & 96.98 \\
\multicolumn{1}{l|}{CycleGAN$^{*}$}  & Attention U-Net\cite{attunet}   & 95.06 & 97.47 & 95.12 & 97.50 & 95.07 & 97.39 & 94.97 & 97.33 & 94.39 & 97.04 \\
\multicolumn{1}{l|}{CycleGAN$^{*}$}  & MedSAM\cite{MedSAM}            & 95.33 & 97.61 & 95.21 & 97.55 & 95.36 & 97.55 & 95.11 & 97.41 & 94.82 & 97.27 \\
\hline
\multicolumn{2}{c|}{BioVessel-Net}                     & \textbf{99.41} & \textbf{99.71} & \textbf{98.66} & \textbf{99.33} & \textbf{99.21} & \textbf{99.60} & \textbf{99.42} & \textbf{99.71} & \textbf{99.19} & \textbf{99.59}  \\
\hline
\multicolumn{12}{l}{* means the model has been optimized.}
\end{tabular}
\label{er-ab}
\end{center}
\end{table*}

\subsubsection*{Hyperparameter Sensitivity Analysis}

\begin{figure}[ht]
  \centering
      \begin{subfigure}{0.22\linewidth}
    \centerline{\includegraphics[width=\columnwidth]{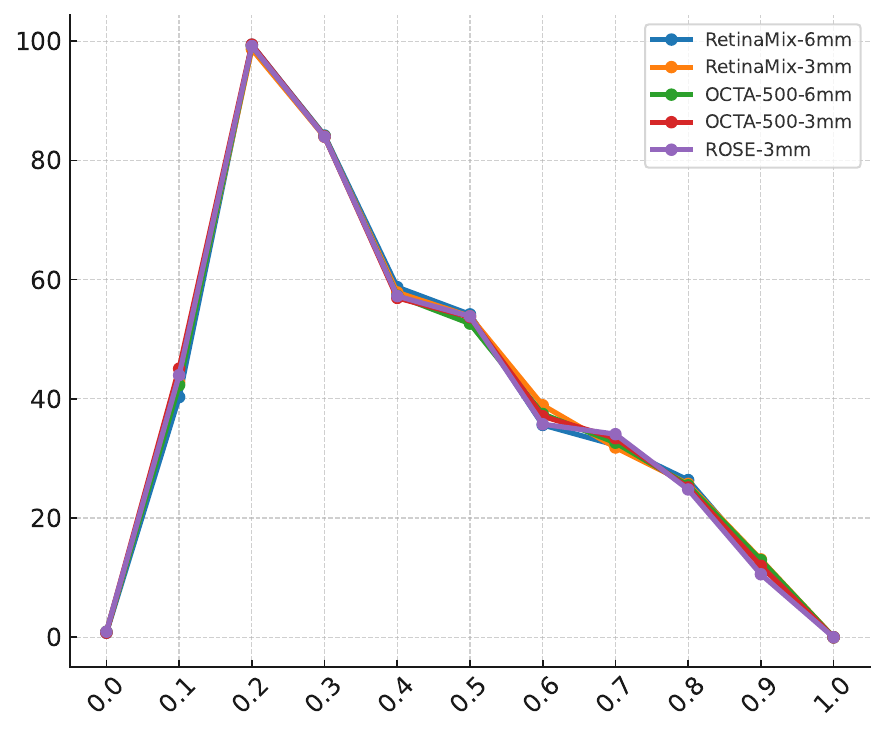}}
    \caption{Impact of $R_{min}$ on IoU}
    \label{ab-a}
  \end{subfigure}
    \hfill
  \begin{subfigure}{0.22\linewidth}
\centerline{\includegraphics[width=\columnwidth]{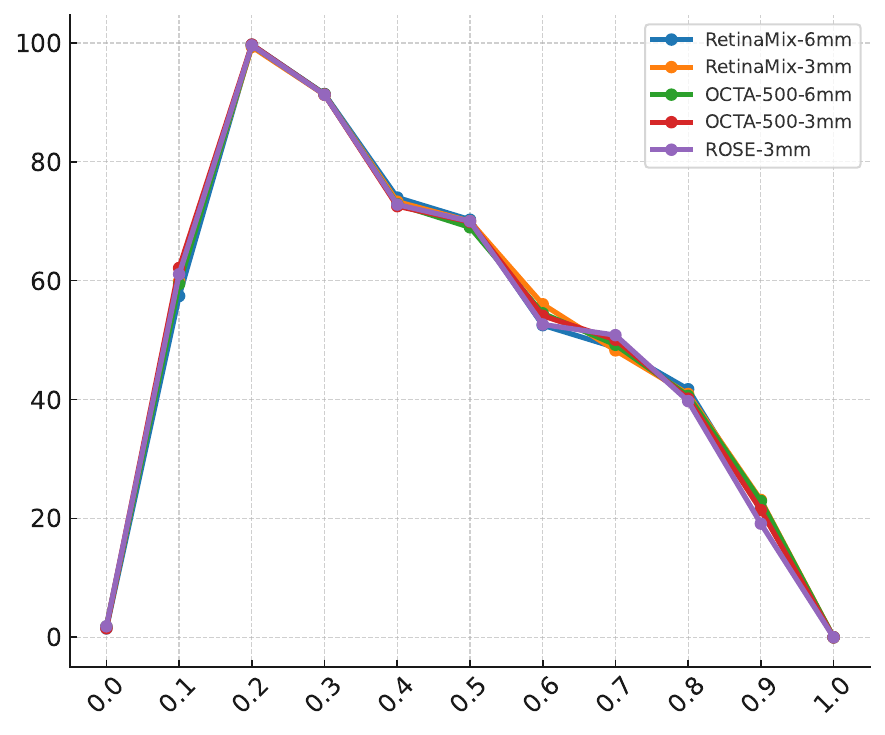}}
    \caption{Impact of $R_{min}$ on Dice}
    \label{ab-b}
  \end{subfigure}
  \hfill
    \begin{subfigure}{0.22\linewidth}
\centerline{\includegraphics[width=\columnwidth]{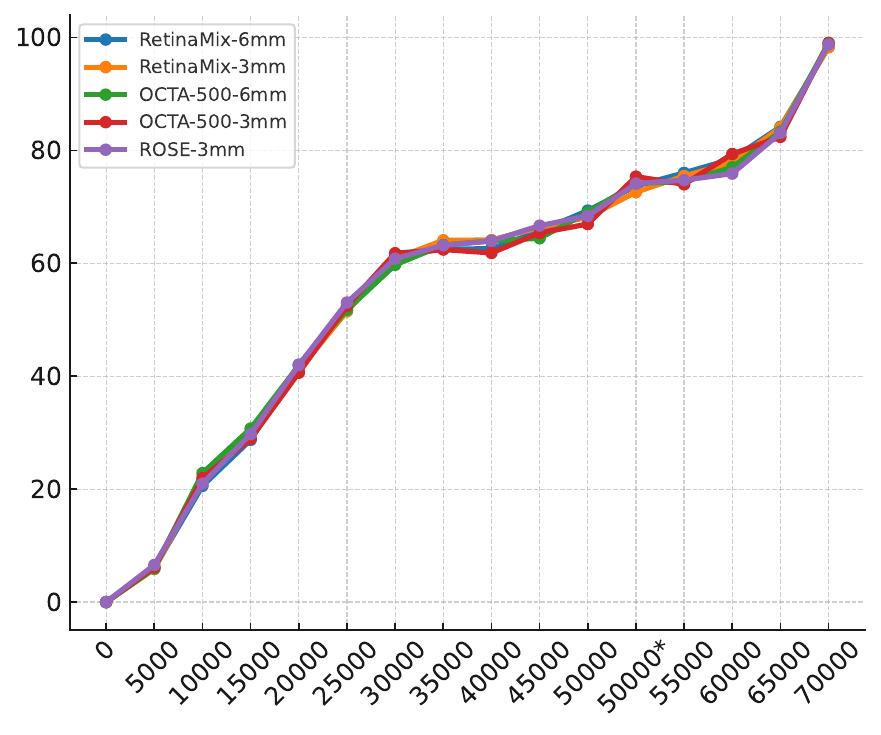}}
    \caption{Impact of Nodes on IoU}
    \label{ab-c}
  \end{subfigure}
  \hfill
  \begin{subfigure}{0.22\linewidth}
\centerline{\includegraphics[width=\columnwidth]{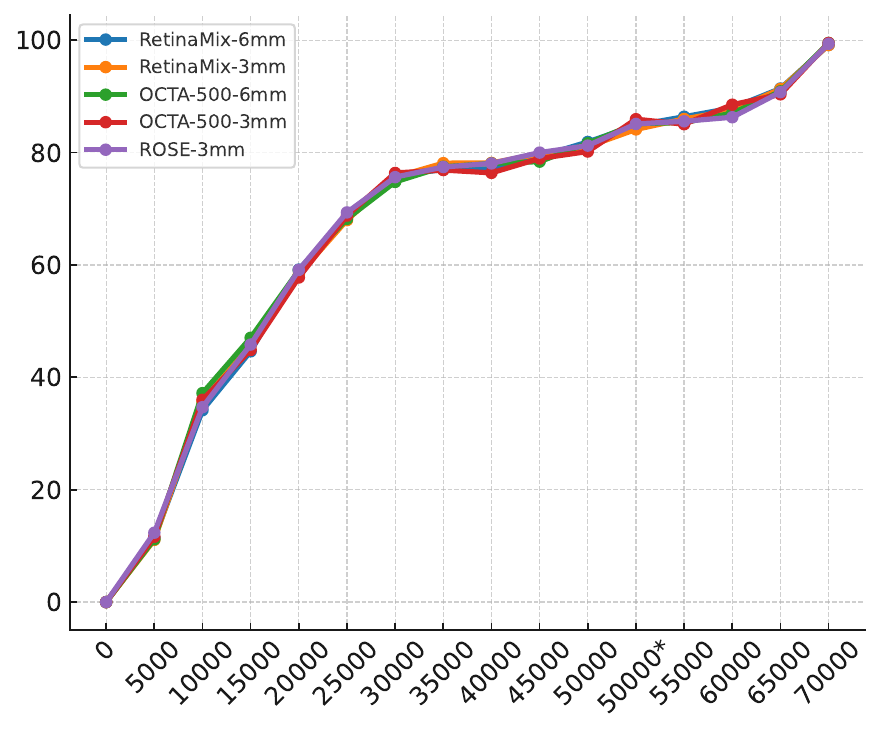}}
    \caption{Impact of Nodes on Dice}
    \label{ab-d}
  \end{subfigure}
  \caption{Ablation Experiment of Hyperparameters ($\times$100\%)}
  \label{ab}
\end{figure}

As shown in Figure~\ref{ab}, the optimal choice of $R_{min}$ (0.2) enables BioVessel-Net to significantly outperform other models in segmentation, which is so high that it is nearly perfect, approaching 100\%. When $R_{min}=1$, BioVessel-Net segments only the largest vessels, resulting in an extremely low segmentation index as only the starting points meet the threshold. When $R_{min}$ is 0, it is equivalent to no filtering, directly the original image. Also, a greater number of generated nodes $N$ leads to smoother vessel edges, a more complete vascular structure, and improved segmentation accuracy. Given that the R values of BioVessel-Net are relatively low, their practical significance indicates minimal differences in the segmented vessels, as the number of nodes remains similar. Consequently, their curves largely overlap, with BioVessel-Net exhibiting a slight increase.

\subsubsection*{Evaluation of Vessel Structure Refinement}

As shown in Table~\ref{ss}, comparison between CycleGAN \cite{cyclegan} and CycleGAN (ours revised version) shows a high structural similarity, with SSIM values ranging from 91.01\% to 96.04\%. CycleGAN (ours) consistently outperforms CycleGAN\cite{cyclegan}, achieving higher SSIM and lower MSE across all datasets. The small SSIM difference (less than 4.31\%) indicates that both models generate highly similar images, but CycleGAN (ours) preserves image structure better and reduces pixel-level errors, making it the superior model.

\begin{table*}[h]
\begin{center}
\caption{Comparison Vessel Structure Refinement Images with Real OCTA Images ($\times$100\%)}
\begin{tabular}{l|cc|cc|cc|cc|cc}
\hline
\multirow{4}{*}{\textbf{Model}}  & \multicolumn{10}{c}{\textbf{Dataset}} \\
\cline{2-11}  &  \multicolumn{4}{c|}{\textbf{RetinaMix}} & \multicolumn{4}{c|}{\textbf{OCTA-500}} & \multicolumn{2}{c}{\textbf{ROSE}}\\
\cline{2-11}  & \multicolumn{2}{c|}{\textbf{6mm}} & \multicolumn{2}{c|}{\textbf{3mm}} & \multicolumn{2}{c|}{\textbf{6mm}} & \multicolumn{2}{c|}{\textbf{3mm}} & \multicolumn{2}{c}{\textbf{3mm}} \\

\cline{2-11}  &  \textbf{SSIM} & \textbf{MSE} & \textbf{SSIM} & \textbf{MSE} & \textbf{SSIM} & \textbf{MSE} & \textbf{SSIM} & \textbf{MSE} & \textbf{SSIM} & \textbf{MSE} \\
\hline
GAN\cite{gan} & 91.76 & 5.69 & 90.39 & 4.43 & 89.69 & 9.85 & 91.06 & 7.36 & 87.84 & 8.73 \\
CycleGAN\cite{cyclegan}   & 93.75 & 2.36 & 94.32 & 2.39 & 93.56 & 1.62 & 94.58 & 4.46 & 91.01 & 3.83 \\
\hline
GAN$^{*}$ & 93.62 & 3.29 & 93.87 & 3.79 & 91.28 & 6.98 & 93.19 & 5.54 & 91.73 & 4.18 \\
CycleGAN$^{*}$ & 95.21 & 1.88 & 95.89 & 1.85 & 94.82 & 1.73 & 96.04 & 3.10 & 95.32 & 2.16 \\
\hline
\multicolumn{11}{l}{* means the model has been optimized.}
\end{tabular}
\label{ss}
\end{center}
\end{table*}

\subsection*{Efficiency Evaluation}

\subsubsection*{Assessment of BioVessel-Net in Terms of Efficiency and Explainability}

As shown in Table~\ref{cc}, conventional SLSMs and G-AI+SLSMs exhibit strong dependence on GPU resources, extensive manual annotations, and computationally intensive training, while lacking explainability and robustness across 2D and 3D imaging scenarios. In contrast, the proposed DFS algorithm eliminates the need for annotations, training, and GPUs, and offers high interpretability owing to its radius-threshold–driven design, though its applicability to generalization and 3D extension remains limited. BioVessel-Net provides a balanced solution by integrating biostatistical priors with generative modeling: it avoids annotation dependency, achieves partial explainability, and demonstrates strong cross-dataset generalization as well as direct scalability to 3D imaging. These results highlight the complementary advantages of DFS in efficiency and interpretability and of BioVessel-Net in clinical applicability and translational potential.

\begin{table}[h]
\begin{center}
\caption{Comparison of Performance Indicators }
\begin{tabular}{l|c|c|c|c}
\hline
\textbf{Indicator}& G-AI+SLSMs & SLSMs & BioVessel-Net & DFS (ours)     \\
\hline
GPU   & $\checkmark$ & $\checkmark$ & $\checkmark$  & $\times$  \\
Annotation   & $\checkmark$/$\times$ & $\checkmark$  & $\times$ & $\times$  \\
Training   &  $\checkmark$ &  $\checkmark$ &  $\checkmark$ & $\times$  \\
Explainability   & $\times$ & $\times$ & $\checkmark$/$\times$  &  $\checkmark$ \\
2D and 3D   & $\times$ & $\times$ & $\checkmark$  &  $\checkmark$ \\
Generalization  & $\checkmark$ & $\times$ & $\checkmark$ & $\times$ \\
\hline
\end{tabular}
\label{cc}
\end{center}
\end{table}

\subsubsection*{Efficiency Analysis of DFS Compared with SLSMs}

Compared to the segmentor, our adjusted DFS offers unmatched advantages, as shown in Table~\ref{ccd}. It requires no separate training or labeling, as DFS is a parameter-efficient algorithm. By eliminating manual annotation, it removes subjective bias from doctors. Additionally, its radius-based segmentation avoids pixel mapping errors in SLSMs, such as interference at vessel edges, inaccuracies at junctions, and mask coverage errors. What is more, the key issue is that both SLSMs and G-AI+SLSMS frameworks are susceptible to fitting problems in segmentation tasks. This arises from the training limitations of the SLSMs model and the nature of the data. In contrast, DFS, as an algorithm rather than a model, is unaffected by this issue. 

\begin{table*}[h]
\begin{center}
\caption{Comparison of Performance Details between DFS and SLSMs}
\label{ccd}
\begin{tabular}{l|c|c|c|c}
\hline
\multirow{2}{*}{\textbf{Model}} &   \multicolumn{4}{c}{\textbf{Details}}   \\
\cline{2-5}& \textbf{Inference Time}     & \textbf{Parameters} & \textbf{FLOPs} & \textbf{GPU Utilization} \\
\hline
U-Net\cite{unet}  & 21ms & 31M & 98 & low\\
U-Net++\cite{unetpp}  & 62ms & 76M & 304 & high\\
Attention U-Net\cite{attunet}   & 83ms & 62M & 215 & slightly high\\
MedSAM\cite{MedSAM}    & 425ms & 109M & 1250 & extremely high\\
\hline
DFS (CPU)    & 0.018ms   & 0M & 0.00028 & -\\
DFS (GPU)    & $\leq$ 0.01ms   & 0M & 0.00028 & extremely low\\
\hline
\end{tabular}
\end{center}
\end{table*}

\subsection*{Visualization}
\begin{figure*}[ht]
  \centering
  \begin{subfigure}{1.0\linewidth}
    \centerline{\includegraphics[width=\columnwidth]{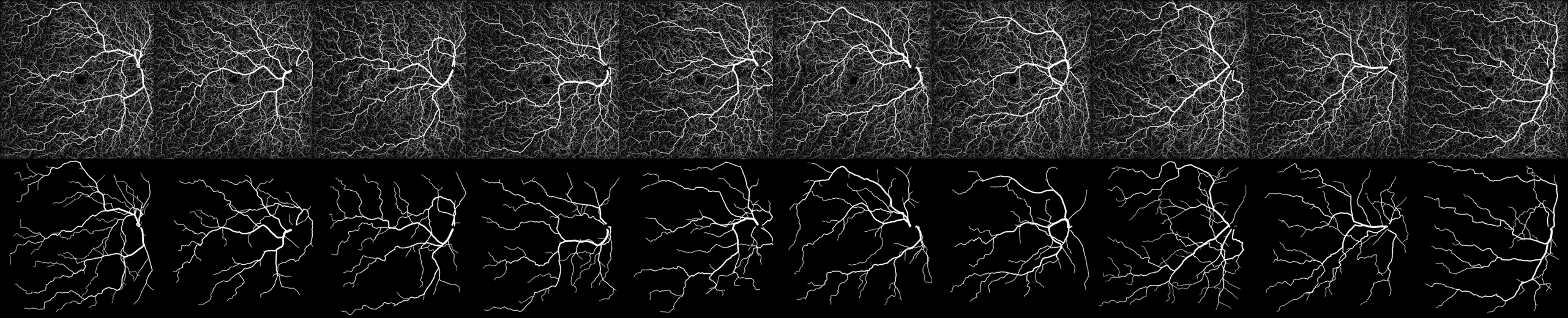}}
    \caption{Segmentation of Main Retinal Vessels (2D)}
    \label{er2d}
  \end{subfigure}
  \hfill
  \begin{subfigure}{0.18\linewidth}
    \centerline{\includegraphics[width=\columnwidth]{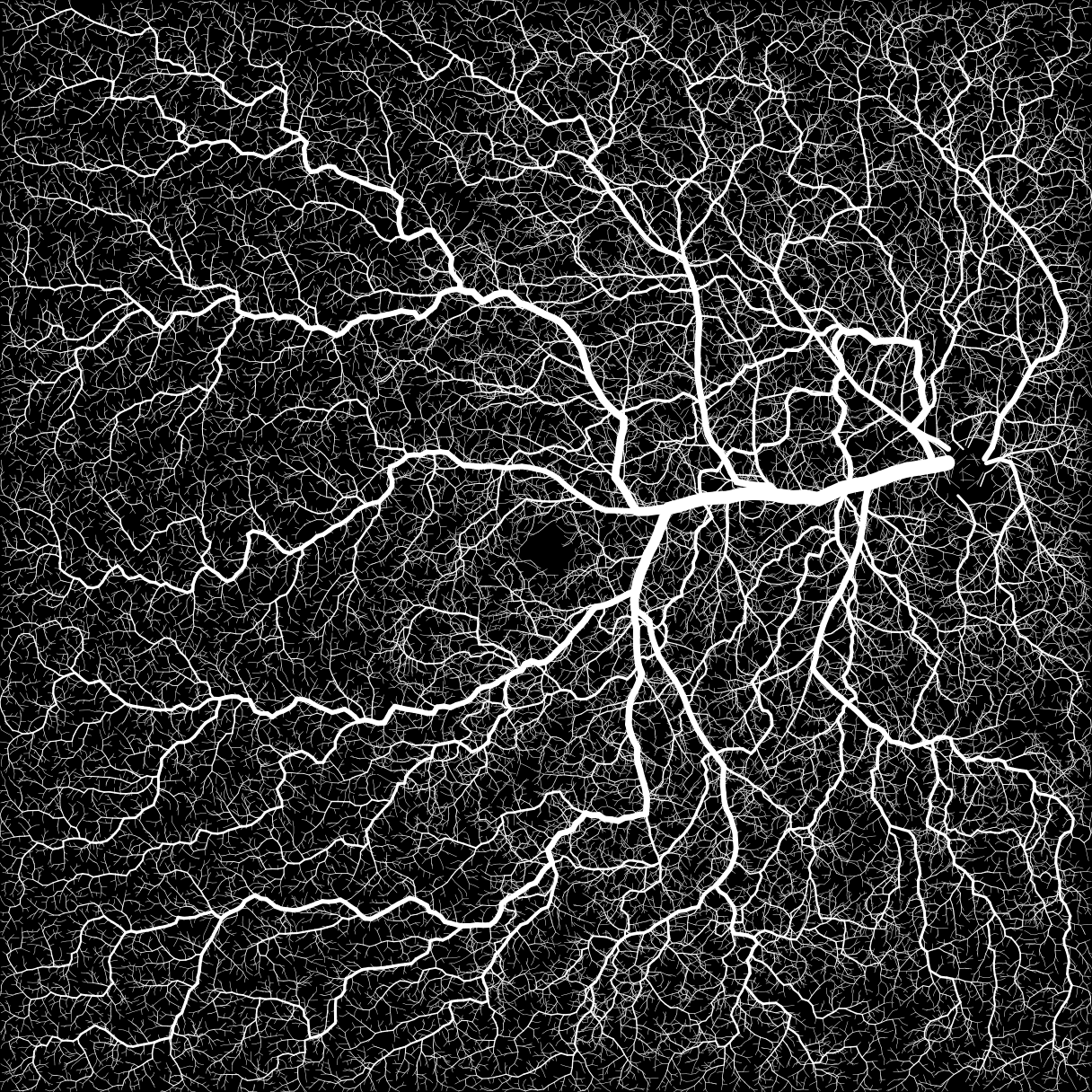}}
    \caption{Original Image (2D)}
    \label{e3d-1}
  \end{subfigure}
    \hfill
  \begin{subfigure}{0.28\linewidth}
    \centerline{\includegraphics[width=\columnwidth]{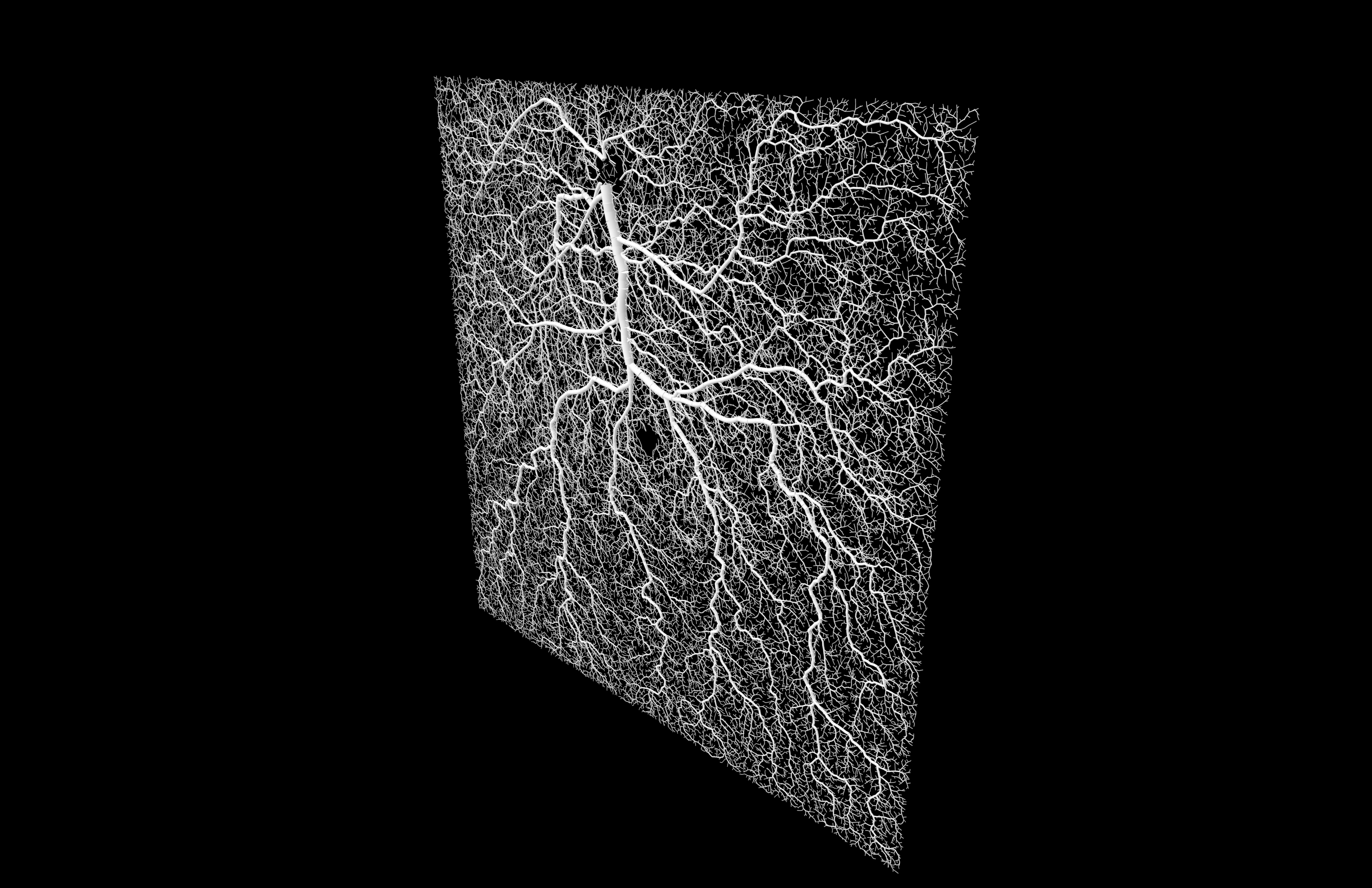}}
    \caption{Original Image (3D)}
    \label{e3d-2}
  \end{subfigure}
      \hfill
  \begin{subfigure}{0.18\linewidth}
    \centerline{\includegraphics[width=\columnwidth]{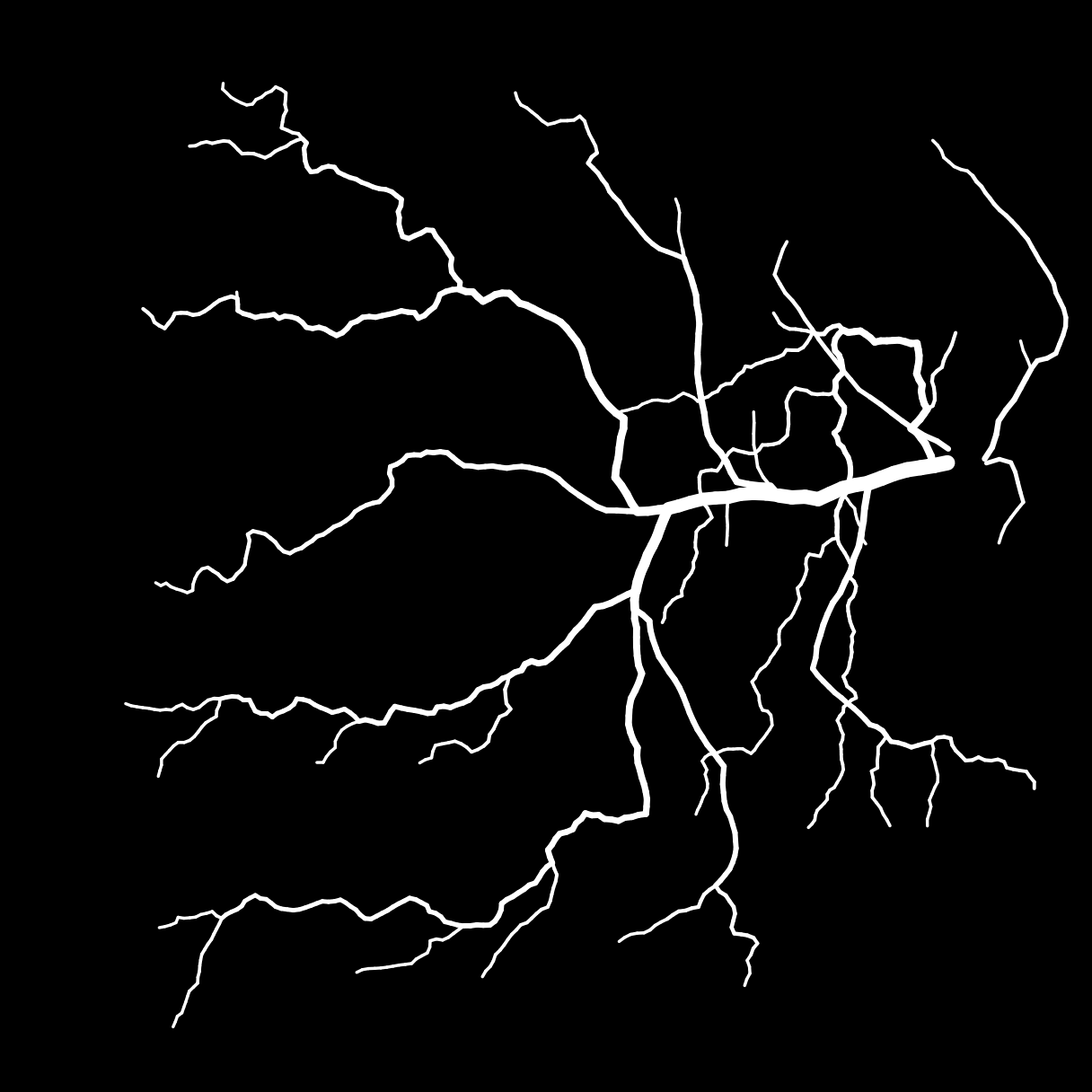}}
    \caption{Segmentation (2D)}
    \label{e3d-3}
  \end{subfigure}
      \hfill
  \begin{subfigure}{0.28\linewidth}
    \centerline{\includegraphics[width=\columnwidth]{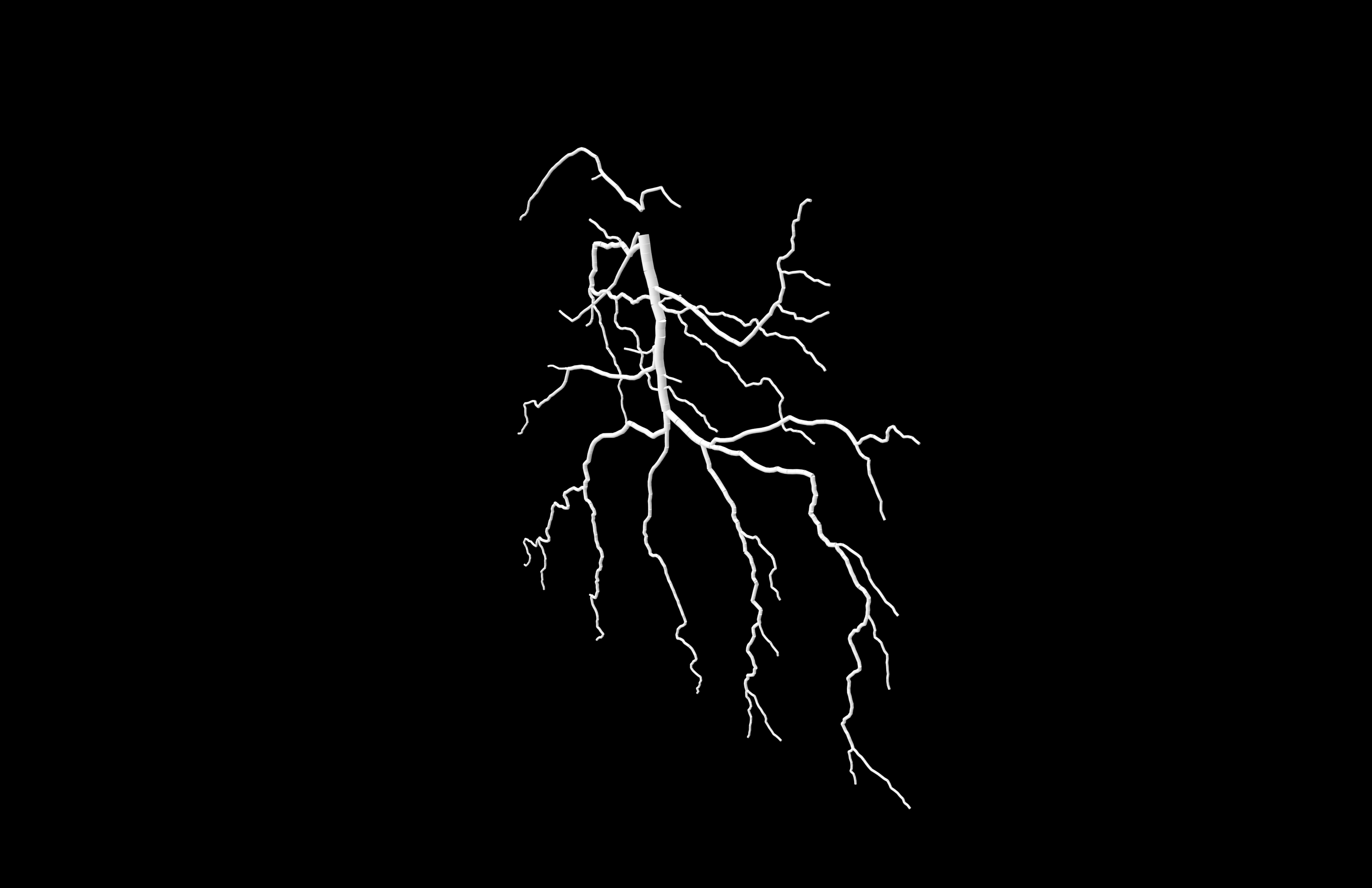}}
    \caption{Segmentation (3D)}
    \label{e3d-4}
  \end{subfigure}
  \caption{Segmentation Results of BioVessel-Net on RetinaMix (2D and 3D)} 
  \label{er-23}
\end{figure*}

As shown in Figure~\ref{er-23}, 2D and 3D segmentation results exhibit smooth, well-defined vessel curves, confirming our prior analysis. In Figure~\ref{er2d}, the segmented vessels are exceptionally smooth, free of pixel artifacts or overflow. DFS effectively filters out vessels below the threshold, eliminating noise and misclassified capillaries. Extending this to 3D, we achieve precise main vessel extraction, with Figure~\ref{e3d-4} preserving vessel integrity and offering a clearer structure than the original image (Figure~\ref{e3d-2}).

\subsection*{Analysis}

An important aspect of our framework is the selection of the radius threshold $R_{min}$, which determines the separation between primary vessels and capillaries. Beyond algorithmic considerations, this parameter exhibits clear biological plausibility. Numerous ophthalmology studies have quantified capillary coverage or vessel density across different retinal plexuses using OCTA and histologic imaging. For instance, Lavia et al.\ \cite{bg-1,bg-3} reported capillary coverage of approximately $30\%$ in the superficial plexus and $17\%$ in the deep plexus, while Xu et al.\ (2024)\cite{bg-2} documented an outer capillary plexus density of $21.9\% \pm 5.9\%$. These values consistently fall within the range of $18$–$26\%$, supporting the validity of adopting $R_{min} \approx 0.2$ as a conservative threshold. Thus, the observed robustness of BioVessel-Net is not only algorithmically sound but also biologically interpretable, reflecting established vascular statistics in the ophthalmic literature.

\section*{Conclusion}
This paper introduced BioVessel-Net, a biologically guided deep learning architecture for accurate segmentation of retinal vessels, and RetinaMix, a high-resolution dataset that provides both two-dimensional and three-dimensional vascular structures with enhanced boundary sharpness and detailed capillary representation. Comprehensive experiments demonstrated that BioVessel-Net achieves state-of-the-art accuracy while preserving anatomical consistency and clinical interpretability. RetinaMix further establishes a superior benchmark for vessel segmentation, enabling rigorous evaluation and fostering future developments in glaucoma-related imaging research. Beyond methodological and dataset contributions, this work highlights the importance of integrating biological priors into model design and curating datasets that reflect fine-grained vascular morphology. These elements together provide a pathway toward more reliable, interpretable, and clinically relevant tools for glaucoma screening and ophthalmic disease analysis.

\section*{Limitation \& Future Work}

While BioVessel-Net demonstrates near-perfect performance across multiple datasets, several limitations remain. 
First, our current evaluation focuses on main vessel segmentation; capturing the full capillary network, especially in diseased eyes with irregular morphology, remains highly challenging. 
Second, although RetinaMix provides a diverse 2D/3D OCTA benchmark, the dataset size is still limited compared to population-scale biobanks, and its generalizability across different imaging devices and clinical settings requires further validation. 
Third, the biological justification of the radius threshold ($R_{min}\!\approx\!0.2$) is supported by existing ophthalmic studies, yet additional large-scale clinical verification is needed to confirm its universality. 
Finally, while the unsupervised design reduces dependency on labeled data, full clinical deployment will require integration with diagnostic pipelines and expert-in-the-loop verification.

Future work will extend BioVessel-Net to fine-grained capillary-level analysis, test its robustness on multi-center OCTA cohorts, and explore hybrid frameworks combining unsupervised biostatistical priors with transformer-based architectures for improved scalability. 
We also plan to release RetinaMix with detailed metadata and baseline benchmarks to facilitate reproducibility and further research in glaucoma imaging.

\section*{Acknowledgements}

This work was supported in part by the National Natural Science Foundation of China under Grants 62276055 and 62406062, in part by the Sichuan Science and Technology Program under Grant 2023YFG0288, in part by the Natural Science Foundation of Sichuan Province under Grant 2024NSFSC1476, in part by the National Science and Technology Major Project under Grant 2022ZD0116100, in part by the Sichuan Provincial Major Science and Technology Project under Grant 2024ZDZX0012.

\section*{Author Contributions Statement}

Cheng Huang developed the model and collected the data. Weizheng Xie performed the experimental validation and implemented the analysis scripts. Fan Gao, Yutong Liu, Ruolng Wu, and Hao Wang were responsible for cleaning the raw data. Zeyu Han and Jingxi Qiu set up and tested the model environment. Dr. Xiangxiang Wang, Dr. Yongbin Yu, and Dr. Zhenglin Wang provided funding support and supervised the overall project.

\bibliography{sample}

\end{document}